\documentclass[10pt,twocolumn,letterpaper]{article}

\usepackage{iccv}
\usepackage{times}
\usepackage{epsfig}
\usepackage{graphicx}
\usepackage{amsmath}
\usepackage{amssymb}

\usepackage{booktabs}
\usepackage{wrapfig}
\usepackage{xspace}
\usepackage{xcolor}
\usepackage{color, colortbl}
\usepackage{subcaption}
\usepackage{multirow}
\usepackage{microtype}
\usepackage{pifont}
\usepackage[accsupp]{axessibility} 
\newcommand{\cmark}{\ding{51}}%
\newcommand{\xmark}{\ding{55}}%
\newcommand{\parsection}[1]{\noindent\textbf{#1 }}

\newcommand*{\ourbenchmark}{VTD\@\xspace}
\newcommand*{\ourbenchmarkname}{Video Task Decathlon\@\xspace}
\newcommand*{\ourbenchmarkfull}{\ourbenchmarkname (\ourbenchmark)\@\xspace}
\newcommand*{\ourmodel}{VTDNet\@\xspace}
\newcommand*{\ourmodelfull}{VTDNet\@\xspace}
\newcommand*{\ourmodelbase}{\hbox{VTDNet}\@\xspace}
\newcommand*{\ourmetric}{VTDA\@\xspace}
\newcommand*{\ourlearn}{CPF\@\xspace}
\definecolor{ao(english)}{rgb}{0.0, 0.5, 0.0}
\definecolor{ceruleanblue}{rgb}{0.16, 0.32, 0.75}
\definecolor{amber}{rgb}{1.0, 0.65, 0.0}

\usepackage[pagebackref=true,breaklinks=true,letterpaper=true,colorlinks,bookmarks=false]{hyperref}

\iccvfinalcopy 

\def\iccvPaperID{2347} 

\begin{document}

\title{\ourbenchmarkname: Unifying Image and Video Tasks \\ in Autonomous Driving}

\author{Thomas E. Huang$^{1}$
\and
Yifan Liu$^{1}$
\and
Luc Van Gool$^{1,2}$
\and
Fisher Yu$^{1}$
\and
$^1$ETH Z\"urich \quad $^2$KU Leuven
\\
\\
\url{https://www.vis.xyz/pub/vtd}
}

\maketitle

\begin{abstract}

Performing multiple heterogeneous visual tasks in dynamic scenes is a hallmark of human perception capability.
Despite remarkable progress in image and video recognition via representation learning,
current research still focuses on designing specialized networks for singular, homogeneous, or simple combination of tasks.
We instead explore the construction of a unified model for major image and video recognition tasks in autonomous driving with diverse input and output structures.
To enable such an investigation, we design a new challenge, \ourbenchmarkfull,
which includes ten representative image and video tasks spanning classification, segmentation, localization, and association of objects and pixels.
On \ourbenchmark, we develop our unified network, \ourmodelbase, that uses a single structure and a single set of weights for all ten tasks.
VTDNet groups similar tasks and employs task interaction stages to exchange information within and between task groups.
Given the impracticality of labeling all tasks on all frames and the performance degradation associated with joint training of many tasks, we design a Curriculum training, Pseudo-labeling, and Fine-tuning (CPF) scheme to successfully train VTDNet on all tasks and mitigate performance loss.
Armed with CPF, \ourmodelbase significantly outperforms its single-task counterparts on most tasks with only 20\% overall computations.
VTD is a promising new direction for exploring the unification of perception tasks in autonomous driving.

\end{abstract}

\section{Introduction}

\begin{figure}[]
\centering
\includegraphics[width=0.98\linewidth]{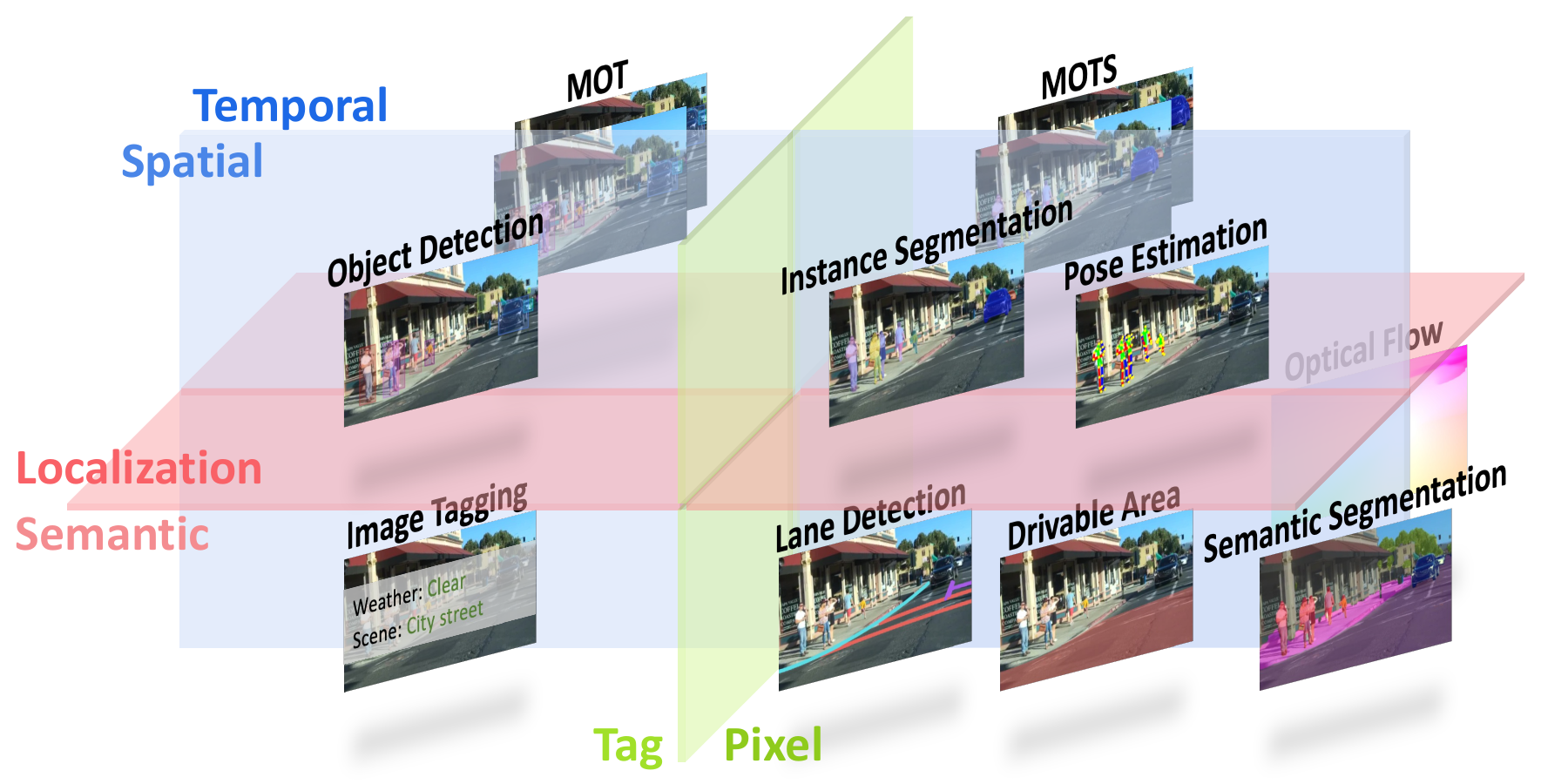}
\vspace{-0.12in}
\caption{
Task categorization of representative image and video recognition tasks.
We design a new challenge and a new architecture to learn a unified representation of image and video tasks for autonomous driving.
}
\label{fig:task-cube}
\vspace{-0.16in}
\end{figure}

Agents that operate in dynamic environments are required to perform a wide range of visual tasks of varying complexities to carry out their functions. For instance, autonomous driving vehicles must identify drivable areas~\cite{wu2021yolop}, detect pedestrians~\cite{milan2016mot16, dendorfer2020mot20}, and track other vehicles~\cite{sun2020waymo, yu2020bdd100k}, among others. Taking a continuous stream of visual inputs, they must be capable of performing tasks at the level of images, instances, and instances across the spatial and temporal extent of the input data. 
While humans can effortlessly complete diverse visual tasks, and representation learning has shown impressive results on individual tasks~\cite{liu2021swin}, there is still a lack of unified architectures that can combine various heterogeneous tasks.

Unified representations for image and video tasks offer numerous advantages, including significant computational savings over using separate networks for each task~\cite{caruana1997multitask}. Additionally, shared task input and output structures~\cite{ye2022inverted, xu2022mtformer} and cascaded tasks~\cite{pang2021quasi, ke2021prototypical} provide opportunities for learning algorithms to exploit inter-task relationships, resulting in better representations, generalization, and overall accuracy~\cite{ruder2017overview}. However, realizing these benefits poses unique challenges. Network architectures must support the predictions of all heterogeneous tasks, which is non-trivial due to the diversity in input and output structures and granularity of visual representation needed for each task.
Furthermore, the impracticality of annotating all video frames for all tasks~\cite{yu2020bdd100k} results in data imbalance between each task and necessitates a more sophisticated training strategy than with single-task or homogeneous multi-task learning.

\begin{figure*}[t]
\centering
\includegraphics[width=\linewidth]{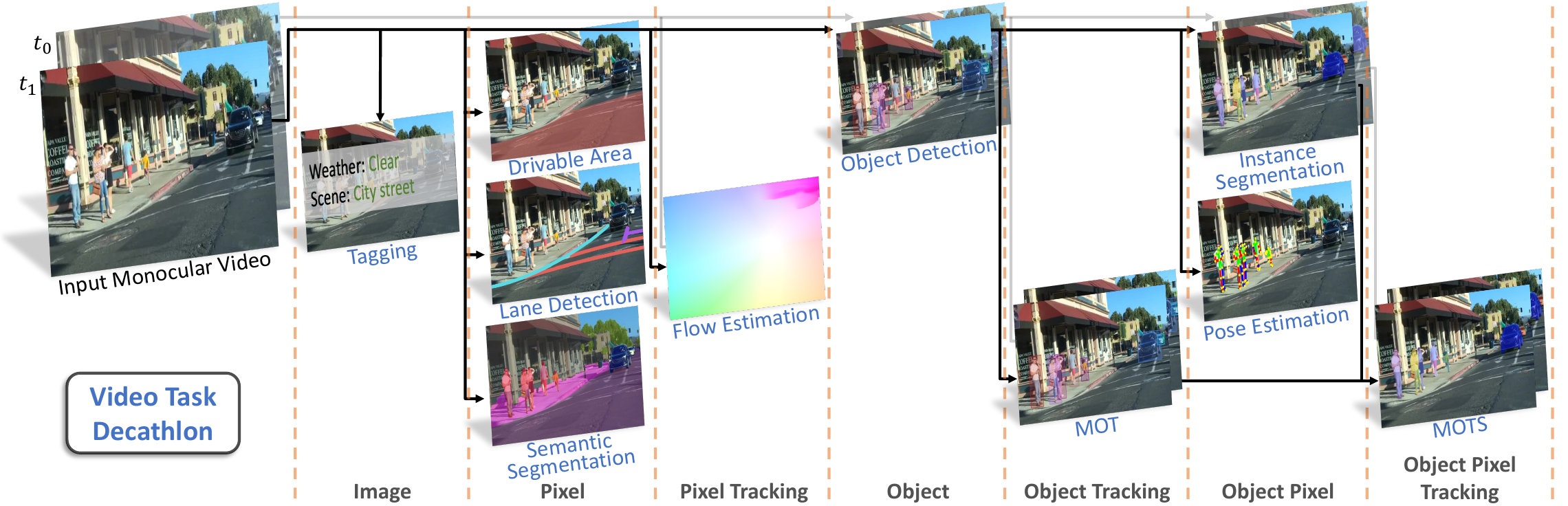}
\vspace{-0.30in}
\caption{
\textbf{\ourbenchmarkfull}. We introduce the \ourbenchmark to study unified representation learning of heterogeneous tasks in 2D vision for autonomous driving.
Given a monocular video, the network needs to produce predictions for ten diverse image and video recognition tasks.
}
\label{fig:task-hierarchy}
\vspace{-0.12in}
\end{figure*}

Another major obstacle to arrive at such a unified representation framework is the lack of large-scale evaluation protocols for heterogeneous combinations of multiple tasks with distinct characteristics.
Current multi-task benchmarks are overly simplistic and focus on combinations of multiple homogeneous tasks such as different types of classifications~\cite{rebuffi2017learning} or pixel-level predictions~\cite{Silberman:ECCV12, mottaghi_cvpr14, Cordts2016Cityscapes, zamir2018taskonomy}.
Those that branch out to different tasks often only consider a limited number of tasks~\cite{nuscenes2019, sun2020waymo, Goel_2021_WACV, Schon_2021_ICCV, liang2022effective}.
In addition, all these works are based solely on image tasks, disregarding the dynamics and associations in videos~\cite{yuan2021florence}.
Although these benchmarks are useful for studying the abstract problem of multi-task learning, they do not adequately support the learning of general representations for the complex, real-world environments encountered in autonomous driving.

To address the aforementioned limitations, we first introduce a new challenge, \textbf{\ourbenchmarkfull}, to study unified representation learning for heterogeneous tasks in autonomous driving.
\ourbenchmark comprises ten visual tasks, chosen to be representative of image and video recognition tasks (\figureautorefname~\ref{fig:task-cube}).
\ourbenchmark provides an all-around test of \textit{classification}, \textit{segmentation}, \textit{localization}, and \textit{association} of objects and pixels.
These tasks have diverse output structures and interdependencies,
making it much more challenging than existing multi-task benchmarks.
Additionally, differences in annotation density between tasks complicate optimization, reflecting real-world challenges.
Along with our challenge, we also propose a new metric, \ourbenchmark Accuracy (\ourmetric),
that is robust to differing metric sensitivities and enables better analysis
in the heterogeneous setting.

To explore unified representation learning on \ourbenchmark, we propose two components: (1) \textbf{\ourmodelfull}, a network capable of training on and producing outputs for every \ourbenchmark task with a \textit{single structure} and a \textit{single set of weights}, and (2) \textbf{\ourlearn}, a progressive learning scheme for joint learning on \ourbenchmark. Specifically, \ourmodel identifies three levels of visual features that are essential for visual tasks, namely image features, pixel features, and instance features.
Each task can be broken down into a combination of these three basic features, and tasks are grouped based on the required features for prediction. Furthermore, \ourmodelbase utilizes \textbf{Intra-group} and \textbf{Cross-group Interaction Blocks} to model feature interactions and promote feature sharing within and across different groups of tasks. 
\ourlearn has three key features:
\textbf{C}urriculum training pre-trains components of the network before joint optimization,
\textbf{P}seudo-labels avoid forgetting tasks without sufficient annotations,
and task-wise \textbf{F}ine-tuning boosts the task accuracies further based on the learned shared representations.
\ourlearn enables \ourmodelbase to jointly learn all the tasks and mitigate a loss of performance.

We conduct experiments for the proposed \ourbenchmark challenge on the large-scale autonomous driving dataset BDD100K~\cite{yu2020bdd100k}.
Armed with \ourlearn, \ourmodelbase is able to significantly outperform strong baselines and other multi-task models on a majority of the tasks and achieve competitive performance on the rest, despite using a single set of weights and only 20\% overall computations.
Our findings indicate that unifying a diverse set of perception tasks for autonomous driving holds great promise for improving performance by leveraging shared knowledge and task relationships, while also achieving greater computational efficiency.

\section{Related Work}

\begin{table*}[t]
\small
\centering
\caption{
Statistics of tasks and available annotations in BDD100K~\cite{yu2020bdd100k}.
}
\label{tab:bdd-splits}
\vspace{-0.14in}
\begin{tabular}{l|cc|c}
\toprule
Set & Images (Train / Val) & \% Total Images & Tasks with Annotations \\ \midrule
Detection & 70K / 10K & 20\% & Tagging, detection, pose, drivable area, lane detection \\
Segmentation & 6.5K / 1K & 2\% & Instance segmentation, semantic segmentation \\
Tracking & 280K / 40K (=1.4K / 200 videos) & 78\% & MOT, MOTS (partially annotated, 31K / 6.4K images) \\ \bottomrule
\end{tabular}
\vspace{-0.12in}
\end{table*}

\parsection{Multi-Task Learning.}
Multi-task learning (MTL)~\cite{caruana1997multitask} is the study of jointly learning and predicting several tasks.
MTL may lead to performance gains due to knowledge sharing between different tasks and better generalization~\cite{ruder2017overview}, while reducing the memory footprint and computational load.
There are two main branches of research: architecture and optimization.
With regard to architecture, early works studied joint learning of pairs of tasks, such as detection and segmentation~\cite{he2017mask}.
Recent works focused on designing models that can learn shared representations from multiple different tasks~\cite{kokkinos2017ubernet, Xu_2018_CVPR, zhang2019pattern, vandenhende2020mti, yuan2021florence, liang2022effective, kanakis2022composite}, including Transformer~\cite{vaswani2017attention} networks~\cite{ye2022inverted, xu2022mtformer, zhang2023demt, ye2023taskprompter}.
Some study networks that can adaptively determine which parameters to share or use for tasks~\cite{maninis2019attentive, kanakis2020reparameterizing}, such as learning a layer-selection policy~\cite{sun2020adashare}.
Others consider utilizing pseudo-labels~\cite{xie2020self, ghiasi2021multi, kanakis2023composite}.
In terms of optimization, most works focus on developing methods to automatically balance the losses of each task~\cite{sener2018multi, chen2018gradnorm, kendall2018multi, yu2020gradient, liu2021towards}.
Some investigate better training strategies, such as task prioritization~\cite{Guo_2018_ECCV}.
In this work, we extend the architectural study to the scale of ten heterogeneous image and video tasks for autonomous driving.

\parsection{Multi-Task Benchmarks.}
Existing MTL datasets typically contain homogeneous image-based tasks and focus on either image classification~\cite{rebuffi2017learning} or dense prediction tasks~\cite{Silberman:ECCV12, mottaghi_cvpr14, Cordts2016Cityscapes, zamir2018taskonomy}.
Recently, new datasets have been proposed to study the combination of object detection, monocular depth estimation, and panoptic segmentation~\cite{Goel_2021_WACV, Schon_2021_ICCV}.
There are also large-scale autonomous driving datasets, mainly for studying detection and tracking~\cite{Geiger2012CVPR, nuscenes2019, sun2020waymo}.
Additionally, synthetic datasets have been developed for autonomous driving~\cite{saleh2018effective, sun2022shift}.
Although these datasets provide a good foundation for the study of MTL, we argue they are either limited in scale or diversity of tasks.
Conversely, BDD100K~\cite{yu2020bdd100k} is a large-scale autonomous driving dataset that contains labels for a diverse set of heterogeneous visual tasks that includes video tasks as well, which enables a new avenue for investigation.
We design a new heterogeneous multi-task challenge based on BDD100K with ten diverse tasks
to enable investigation of the unification of image and video tasks for autonomous driving.

\section{\ourbenchmarkname}
We introduce \ourbenchmarkfull, a new challenge for investigating heterogeneous multi-task learning on a diverse set of 2D video tasks for autonomous driving (\figureautorefname~\ref{fig:task-hierarchy}).
The goal is to facilitate designing models capable of handling all 2D tasks on monocular video frames.
\ourbenchmark comprises ten tasks: image tagging, object detection, pose estimation, drivable area segmentation, lane detection, semantic segmentation, instance segmentation, optical flow estimation, and multi-object tracking (MOT) and segmentation (MOTS).
These tasks are representative of the space of 2D vision for autonomous driving (\figureautorefname~\ref{fig:task-cube}).

\parsection{Dataset.} \label{section-bdd100k}
We build \ourbenchmark on top of the real-world large-scale BDD100K video dataset~\cite{yu2020bdd100k}, which has annotations for a diverse range of vision tasks.
BDD100K consists of 100K driving video sequences, each around 40 seconds long.
The tracking tasks are annotated at 5 FPS.
The tasks are annotated on three separate image sets, which are all subsets of the original 100K videos.
However, each image set only has labels for a portion of the available tasks.
The statistics (after data deduplication) and tasks within each set are shown in \tableautorefname~\ref{tab:bdd-splits}.
The varying size of each set reflects real-world difficulties in annotation.
Consequently,
this complicates optimization as different tasks have different data proportions and each image is only partially labeled.

\subsection{Tasks} \label{section-vtd}
In this section, we describe each task in \ourbenchmark.
Due to space constraints, we provide additional task-specific details in \sectionautorefname~\ref{sec:sup-benchmark}.

\parsection{Image Tagging (G).}
Image tagging is composed of two classification tasks aiming
to classify the input image into one of seven different weather conditions and seven scene types.
We use the top-1 accuracy as the metric for each task ($\text{Acc}^{\texttt{Gw}}$ for weather and $\text{Acc}^{\texttt{Gs}}$ for scene).

\parsection{Drivable Area Segmentation (A).}
The drivable area task involves predicting which areas of the image are drivable.
We use the standard mIoU metric ($\text{IoU}^{\texttt{A}}$).

\parsection{Lane Detection (L).}
The task of lane detection is to predict the lane types and their location in the image.
We treat lane detection as a contour detection problem.
For evaluation, we use the boundaries mIoU metric ($\text{IoU}^{\texttt{L}}$).

\parsection{Semantic (S) / Instance Segmentation (I).}
Semantic and instance segmentation involves predicting the category and instance label for every pixel.
We use the mIoU metric for semantic segmentation ($\text{IoU}^{\texttt{S}}$) and the mask AP metric for instance segmentation ($\text{AP}^{\texttt{I}}$).

\begin{figure*}[t]
\centering
\includegraphics[width=\linewidth]{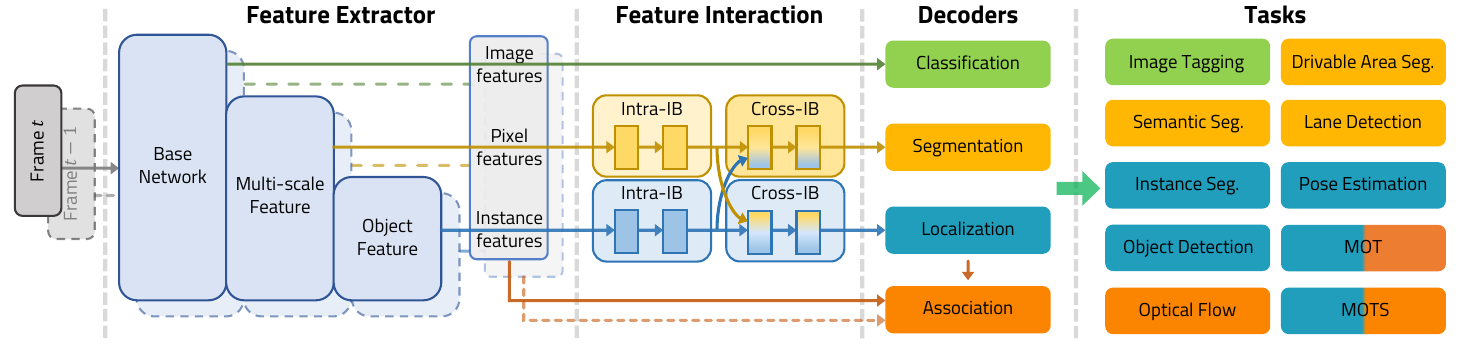}
\vspace{-0.24in}
\caption{
\textbf{Unified architecture of \ourmodel.}
Tasks are grouped into \textcolor{ao(english)}{classification}, \textcolor{amber}{segmentation}, \textcolor{ceruleanblue}{localization}, and \textcolor{orange}{association}.
\ourmodel includes a shared feature extractor to extract hierarchical features, feature interaction blocks to exchange knowledge between tasks, and decoders to make the final prediction for each task.
}
\label{fig:full-model}
\vspace{-0.12in}
\end{figure*}

\parsection{Object Detection (D) / Pose Estimation (P).}
Object detection involves predicting a 2D bounding box and the category for each object in the image.
Pose estimation involves localizing 2D joint positions for each human in the image.
For evaluation, we use the bounding box AP metric ($\text{AP}^{\texttt{D}}$) for detection and Object Keypoint Similarity (OKS) AP metric ($\text{AP}^{\texttt{P}}$) for pose estimation.

\parsection{MOT (T) / MOTS (R).}
MOT involves detecting and tracking every object in the video sequence.
MOTS also includes segmentation of every object.
For evaluation, we use a combination of AP for measuring detection and segmentation performance ($\text{AP}^{\texttt{T}}$ and $\text{AP}^{\texttt{R}}$) and AssA from HOTA~\cite{luiten2021hota} for measuring association performance ($\text{AssA}^{\texttt{T}}$ and $\text{AssA}^{\texttt{R}}$).

\parsection{Optical Flow Estimation (O).}
Optical Flow estimation is the task of determining pixel-wise motions between pairs of images.
As BDD100K does not have labels for optical flow,
we use a proxy evaluation method based on MOTS labels
by warping the segmentation masks with the predicted flow and using the overlap with the ground-truth masks as the score~\cite{hsuan2016objectflow}.
We use mean IoU as the metric ($\text{IoU}^{\texttt{F}}$).

\subsection{Evaluation}
We here introduce our metric for evaluating model performances on \ourbenchmark.
There are two difficulties with designing a metric for a heterogeneous multi-task setup.
First, as different tasks have different metrics, their sensitivity also varies, causing task-wise improvements to differ in scale.
Second, due to the number of tasks and inter-class overlap,
only looking at task-specific metrics will not give a clear indication of the model's overall performance, and a simple average over all metrics will hide task-wise differences.

To address these issues, we propose our VTD Accuracy (VTDA) metric.
We first account for differing metric sensitivities by using the standard deviation in their measurements to scale their score accordingly.
We estimate the standard deviation $\sigma_t$ of each task $t$ by measuring it over single-task baseline performances across different base networks (\sectionautorefname~\ref{sec:main-res}), which informs how each metric's values change across increasing network capacity and differing architectures.
Next, we discretize $\sigma_t$ estimates to account for noise and convert them to a scaling factor by $s_t = 1 / \lceil 2 \sigma_t \rceil$, such that metrics with lower standard deviation will be scaled higher as differences are more significant and vice versa.
This ensures that task scores contribute similarly to the final score and reduces bias towards a particular task.

Additionally, to better analyze multi-task performance, we separate the ten tasks into four groups first, each measuring a key aspect of the network's performance: classification, segmentation, localization, and association.

\parsection{Classification.}
This group includes the two classification tasks in image tagging.

\parsection{Segmentation.}
Segmentation refers to tasks that require prediction of a class label for each pixel in the image, \ie, dense prediction.
In \ourbenchmark, this includes semantic segmentation, drivable area segmentation, and lane detection.

\parsection{Localization.}
Localization includes object detection (bounding box), instance segmentation (pixel mask), and pose estimation (keypoints).
We also consider detection and instance segmentation for object tracking (MOT and MOTS).

\parsection{Association.}
Association includes optical flow estimation (image pixels), MOT (object bounding boxes), and MOTS (object pixel masks).
In this group, we only evaluate the association performance of tracking,
as localization errors are already accounted for in the localization group.

\parsection{VTDA.}
We take the average of the scaled task performance within each group to compute a corresponding measure,
\begin{align}
\begin{split}
\text{VTDA}_{\text{\textcolor{ao(english)}{cls}}} &= (\text{Acc}^{\texttt{Gw}}_{s} + \text{Acc}^{\texttt{Gs}}_{s})/2,\\
\text{VTDA}_{\text{\textcolor{amber}{seg}}} &= \left(\text{IoU}^{\texttt{S}}_{s} + \text{IoU}^{\texttt{A}}_{s} + \text{IoU}^{\texttt{L}}_{s}\right)/3,\\
\text{VTDA}_{\text{\textcolor{ceruleanblue}{loc}}} &= \left(\text{AP}^{\texttt{D}}_{s} + \text{AP}^{\texttt{I}}_{s} + \text{AP}^{\texttt{P}}_{s} + \text{AP}^{\texttt{T}}_{s} + \text{AP}^{\texttt{R}}_{s}\right)/5,\\
\text{VTDA}_{\text{\textcolor{orange}{ass}}} &= \left(\text{AssA}^{\texttt{T}}_{s} + \text{AssA}^{\texttt{R}}_{s} + \text{IoU}^{\texttt{F}}_{s}\right)/3,
\end{split}
\end{align}
where the subscript $s$ denotes scaling.
Each score is normalized to the range [0, 100],
and VTDA is defined as the sum of all scores.
We provide full details about \ourmetric including the scaling factors used in \sectionautorefname~\ref{sec:sup-metric}.

\section{Method}
To tackle \ourbenchmark, we propose \ourmodelfull, a network capable of learning a unified representation for all ten tasks.
We describe the network architecture in detail in \sectionautorefname~\ref{section-model}, feature interaction blocks in \sectionautorefname~\ref{section-featint}, and the optimization protocol in \sectionautorefname~\ref{section-opt}.
The architecture is shown in \figureautorefname~\ref{fig:full-model}.

\begin{figure}[t]
\centering
\includegraphics[width=\linewidth]{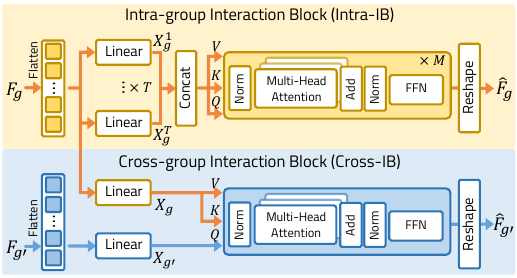}
\vspace{-0.24in}
\caption{
\textbf{Feature interaction blocks in \ourmodel.}
We use self- and cross-attention modules to model feature interactions within and between groups of tasks.
}
\label{fig:attn-blocks}
\vspace{-0.14in}
\end{figure}

\subsection{Heterogeneous Multi-Task Network} \label{section-model}
The heterogeneous nature of the \ourbenchmark tasks necessitates input features to be extracted in a similar hierarchical manner, as different tasks require features at different visual granularities.
\ourmodelbase first uses a shared feature extractor to obtain three levels of visual features, namely image features, pixel features, and instance features.
These features are essential for visual tasks and can be used to tackle the \ourbenchmark tasks.
Additionally, we separate the tasks into four task groups following \ourmetric: classification, segmentation, localization, and association.
Tasks in each group operate at the same feature level and can leverage shared processing to exchange information within the group (\sectionautorefname~\ref{section-featint}).
Independent decoders are used to produce the final predictions for each task.
We detail the feature extractor and each decoder group below.
To save space, we omit further details
and elaborate them in \sectionautorefname~\ref{sec:sup-hdet}.

\parsection{Feature Extractor}
consists of the base network, a multi-scale feature extractor, and an object feature extractor to obtain hierarchical features.
The base network produces image features $\{C2,C3,C4,C5\}$ with strides $\{2^2,2^3,2^4,2^5\}$.
Next, we use a Feature Pyramid Network (FPN)~\cite{lin2017feature} to construct a multi-scale feature pyramid based on the image features
and produce pixel features $\{P2,P3,P4,P5,P6\}$.
Finally, we use a Region Proposal Network (RPN)~\cite{ren2015faster} to produce instance features at each scale.
For simplicity, we use an image-based base network rather than a video-based one, which enables \ourmodelbase to operate online.
For video tasks, we apply the same feature processing to additional video frames independently.

\parsection{Classification Decoders}
operate on image features for prediction.
The image tagging decoder uses global average pooling on $C5$ and has two dense layers, one for each classification task.

\parsection{Segmentation Decoders}
require high resolution pixel features to output per-pixel predictions.
We progressively upsample each FPN feature map to the same scale as $P2$ using convolutions and aggregate them with element-wise summation.
Given the aggregated features, the drivable area, lane detection, and semantic segmentation decoders each use convolutional layers to obtain the final outputs.

\parsection{Localization Decoders}
utilize instance features to make predictions for every object in the image.
The detection, instance segmentation, and pose estimation decoders
use parallel convolutional branches to map the instance features to the desired output~\cite{he2017mask}.

\parsection{Association Decoders}
are built on the previous features and decoder outputs across pairs of video frames.
The flow estimation decoder uses warping on the features from the first two pixel feature maps $P2$ and $P3$ to construct a cost volume and convolutions to predict the flow, following standard procedure~\cite{sun2018pwc}.
The MOT and MOTS decoders associate objects predicted by the detection and instance segmentation decoders using a learned similarity measure through contrastive learning, following QDTrack~\cite{fischer2022qdtrack, pang2021quasi}.

\parsection{Training Loss.}
Our joint training loss function is defined as a linear combination of all losses
$L_\mathtt{VTD} = \sum_t \lambda_t L_t$ for each task $t$ with corresponding loss weight $\lambda_t$.

\subsection{Feature Interaction}  \label{section-featint}
To further enhance knowledge sharing between tasks, we augment \ourmodelbase with explicit pathways to incorporate additional avenues for task interactions and information exchange.
We include such pathways by adding feature interaction blocks between similar tasks in the same group (intra-group) and between tasks in different groups (cross-group).
These blocks are placed after the feature extractor and before the task decoders, and they are shown in \figureautorefname~\ref{fig:attn-blocks}.

\parsection{Intra-group Interaction Block (Intra-IB).}
Similar tasks within the same group can benefit from additional shared processing to model task interactions before independent task decoding.
We incorporate interaction blocks based on attention~\cite{vaswani2017attention} to model such interactions.
Specifically, for a particular task group $g$ and input features $F_g\in\mathbb{R}^{H\times W\times C}$, we first flatten into tokens and use a set of linear layers $L_g^1,\ldots,L_g^{T}$ to extract task-specific tokens ${X_g^1,\ldots,X_g^T\in\mathbb{R}^{HW\times C'}}$, where $T$ is the number of tasks in group $g$.
After concatenation ${\hat{X}_g=\textrm{Concat}\left(X_g^1,\ldots,X_g^T\right)}\in\mathbb{R}^{HW\times TC'}$, we use a series of self-attention blocks for modeling interactions, each of which consists of Layer Normalization (LN)~\cite{ba2016layer}, Multi-Head Attention (MHA), and a feedforward network (FFN):
\begin{equation}
\begin{split}
    Q=\textrm{LN}(&\hat{X}_g^i), K=\textrm{LN}(\hat{X}_g^i), V=\textrm{LN}(\hat{X}_g^i), \\
    \hat{X}_g^{\prime i}&=\textrm{MHA}(Q, K, V)+\hat{X}_g^i, \\
    \hat{X}_g^{i+1}&=\textrm{FFN}(\textrm{LN}(\hat{X}_g^{\prime i}))+\hat{X}_g^{\prime i},
\end{split}
\end{equation}
where $i$ indicates the $i$-th self-attention block and $Q, K, V$ are the query, key, and value matrices.
We use $M$ such self-attention blocks ($M=2$ in our experiments).
Finally, we reshape $\hat{X}_g$ back to the input feature dimensions to obtain output features $\hat{F}_g\in\mathbb{R}^{H\times W\times TC'}$.

We use Intra-IB with the segmentation and localization decoder groups.
Since, Intra-IB introduces more parameters and computation to the task group,
we reduce the size of each decoder in the group to offset the increase in computation, which makes them more lightweight and enables \ourmodelbase to maintain its advantage in efficiency.

\parsection{Cross-group Interaction Block (Cross-IB).}
Tasks in different groups can also benefit from sharing feature representations.
For example, instance features can inject more knowledge regarding foreground objects to the segmentation task group, while pixel features can provide more information regarding background regions for the localization task group.
We integrate additional feature interaction blocks between different task groups.
Specifically, for any two task groups $g$ and $g^\prime$ with corresponding input features $F_g\in\mathbb{R}^{H\times W\times C}$ and $F_{g^\prime}\in\mathbb{R}^{H^\prime\times W^\prime\times C^\prime}$,
we flatten and use a pair of linear layers $L_{g}$ and $L_{g^\prime}$ to obtain task group tokens $X_g\in\mathbb{R}^{HW\times C''}$ and $X_{g^\prime}\in\mathbb{R}^{H^\prime W^\prime \times C''}$.
For interaction, we use a cross-attention module to incorporate information from one task to another:
\begin{equation}
\begin{split}
    Q=\textrm{LN}(&X_{g^\prime}), K=\textrm{LN}(X_g), V=\textrm{LN}(X_g), \\
    \hat{X}^{\prime}_{g^\prime}&=\textrm{MHA}(Q, K, V)+X_{g^\prime}, \\
    \hat{X}_g^\prime&=\textrm{FFN}(\textrm{LN}(\hat{X}^{\prime}_{g^\prime}))+\hat{X}^{\prime}_{g^\prime},
\end{split}
\end{equation}
where the task tokens from one group is used to query the features of the other group.
Finally, we reshape $\hat{X}_g^\prime$ back to the input feature dimensions to obtain output features $\hat{F}_g^\prime\in\mathbb{R}^{H\times W\times C}$.
We place Cross-IB after Intra-IB to model interactions between the segmentation and localization task groups in both directions.

\subsection{Joint Learning} \label{section-opt}
There are two main optimization challenges in \ourbenchmark: diversity in annotation density and in difficulty of tasks.
Different tasks have different trade-offs between annotation variety and density.
For example, detection is labeled on sampled frames from 100K videos, while tracking is labeled on only 2K videos, which has more frames in total but lower variety.
Additionally, different tasks require different numbers of optimization steps to converge.
These problems are further exacerbated by the large number of tasks.
Naive joint optimization of all tasks will lead to significantly worse performance (\sectionautorefname~\ref{sec:main-res}).
To address these challenges, we construct a progressive training protocol called \ourlearn, which has three key features: \textbf{C}urriculum training, \textbf{P}seudo-labeling, and \textbf{F}ine-tuning.

\parsection{Curriculum Training.}
In order to handle the difference in difficulty of tasks,
we follow a curriculum training protocol where we first pre-train \ourmodelbase on a subset of the tasks then jointly train on all tasks.
During pre-training, we train the localization and object tracking decoders on all relevant data, as they require more optimization steps.
This enables the entire feature extractor and data-hungry task decoders (\eg, MOT) to be initialized before joint training, which greatly improves final multi-task performance.
After pre-training, we jointly train our model on all tasks.
We use a set-level round-robin sampling scheme for
data sampling,
which samples a batch from each image set in order.
At each step, only the weights of task decoders that receive corresponding data are updated.
For efficiency, we do not oversample from any image set and cycle through each image only once per epoch.

\parsection{Pseudo-labeling.}
Due to differences in annotation density between tasks,
joint training with all ten tasks will lead to bias in performance towards the label-dominant tasks.
The performance of tasks with a smaller proportion of labels (semantic segmentation and pose estimation in \ourbenchmark) will thus suffer.
To combat this issue, we utilize the single-task baselines to generate pseudo-labels to provide more labels for these tasks during joint training, which mitigates performance loss due to underfitting.
We use the same training loss for pseudo-labels as the original task loss.
As our goal is to address label deficiency, we only generate pseudo-labels for semantic segmentation and pose estimation.

\parsection{Fine-tuning.}
During joint training, most task decoders will only receive a training signal periodically.
This means the input feature distribution will have shifted before the next gradient update, making it difficult for each decoder to fully converge.
Additionally, gradients from other tasks may also interfere with the training.
To alleviate these issues, we further fine-tune each decoder on its task data after joint training while freezing the rest of the network (including shared blocks).
This is akin to downstream fine-tuning with the learned shared representation and
enables each decoder to obtain dedicated training without interference.

\definecolor{LightCyan}{rgb}{0.88,1,1}
\setlength{\tabcolsep}{2.9pt}
\begin{table*}[]
\scriptsize
\centering
\caption{
Comparison of \ourmodelbase against single-task (ST) and multi-task (MT) baselines on VTD validation set.
CPF denotes our training protocol, and $\dagger$~denotes a separate model is trained for each task.
\ourmodelbase outperforms both ST and MT baselines on most tasks across both base networks and achieves significantly better \ourmetric.
\textbf{Black} / \textit{\textcolor{blue}{blue}} indicate best / second best.
}
\label{tab:main-results-backbones}
\vspace{-0.16in}
\begin{tabular*}{\linewidth}{@{\extracolsep{\fill}}lcc|clc|cccc|cccccc|cccc|c}
\toprule
 &  &  & \multicolumn{2}{c}{Classification} &  & \multicolumn{3}{c}{Segmentation} &  & \multicolumn{5}{c}{Localization} &  & \multicolumn{3}{c}{Association} &  &  \\
 &  &  & \multicolumn{2}{c}{Tagging} &  & Sem. & Driv. & Lane &  & Det. & Ins. & Pose & MOT & MOTS &  & Flow & MOT & MOTS &  &  \\
\multirow{-3}{*}{Method} & \multirow{-3}{*}{\begin{tabular}{@{}c@{}}Base\\Network\end{tabular}} & \multirow{-3}{*}{CPF} & $\text{Acc}^{\texttt{Gw}}$ & $\text{Acc}^{\texttt{Gs}}$ & \multirow{-3}{*}{\parbox[t]{1.5mm}{\tiny \rotatebox[origin=c]{90}{VTDA$_{\text{\textcolor{ao(english)}{cls}}}$}}} & $\text{IoU}^{\texttt{S}}$ & $\text{IoU}^{\texttt{A}}$ & $\text{IoU}^{\texttt{L}}$ & \multirow{-3}{*}{\parbox[t]{0mm}{\tiny \rotatebox[origin=c]{90}{VTDA$_{\text{\textcolor{amber}{seg}}}$}}} & $\text{AP}^{\texttt{D}}$ & $\text{AP}^{\texttt{I}}$ & $\text{AP}^{\texttt{P}}$ & $\text{AP}^{\texttt{T}}$ & $\text{AP}^{\texttt{R}}$ & \multirow{-3}{*}{\parbox[t]{0mm}{\tiny \rotatebox[origin=c]{90}{VTDA$_{\text{\textcolor{ceruleanblue}{loc}}}$}}} & $\text{IoU}^{\texttt{F}}$ & $\text{AssA}^{\texttt{T}}$ & $\text{AssA}^{\texttt{R}}$ & \multirow{-3}{*}{\parbox[t]{0mm}{\tiny \rotatebox[origin=c]{90}{VTDA$_{\text{\textcolor{orange}{ass}}}$}}} & \multirow{-3}{*}{VTDA} \\ \midrule
ST Baselines$\dagger$ &  &  & 81.9 & 77.9 & 80.6 & 59.7 & 83.9 & \textbf{28.4} & {\color[HTML]{0000FF} \textit{56.7}} & 32.3 & 20.2 & {\color[HTML]{0000FF} \textit{37.0}} & 32.9 & 27.2 & 29.7 & 59.6 & 48.8 & 42.4 & 51.3 & 218.2 \\
 &  & \textcolor{lightgray}{\xmark} & \textbf{83.5} & 79.2 & \textbf{82.1} & 45.1 & {\color[HTML]{0000FF} \textit{85.2}} & 26.7 & 54.1 & {\color[HTML]{0000FF} \textit{32.7}} & {\color[HTML]{0000FF} \textit{26.5}} & 29.6 & 31.2 & 28.5 & 30.0 & \textbf{60.6} & 46.9 & 41.6 & 50.7 & 216.9 (\textcolor{purple}{-1.3}) \\
\multirow{-2}{*}{MT Baseline} &  & \textcolor{ao(english)}{\cmark} & 83.0 & {\color[HTML]{0000FF} \textit{79.4}} & 81.8 & {\color[HTML]{0000FF} \textit{60.6}} & {\color[HTML]{0000FF} \textit{85.2}} & 25.9 & 56.4 & 32.5 & 26.4 & 35.0 & {\color[HTML]{0000FF} \textit{33.8}} & {\color[HTML]{0000FF} \textit{30.2}} & {\color[HTML]{0000FF} \textit{31.5}} & {\color[HTML]{0000FF} \textit{60.3}} & {\color[HTML]{0000FF} \textit{49.0}} & {\color[HTML]{0000FF} \textit{43.3}} & {\color[HTML]{0000FF} \textit{51.8}} & {\color[HTML]{0000FF} \textit{221.5 (\textcolor{ao(english)}{+3.3})}} \\
\rowcolor{LightCyan} \textbf{\ourmodel} & \multirow{-4}{*}{ResNet-50} & \textcolor{ao(english)}{\cmark} & {\color[HTML]{0000FF} \textit{83.2}} & \textbf{79.7} & {\color[HTML]{0000FF} \textit{82.0}} & \textbf{63.8} & \textbf{85.4} & {\color[HTML]{0000FF} \textit{27.8}} & \textbf{57.8} & \textbf{33.4} & \textbf{27.1} & \textbf{39.7} & \textbf{34.7} & \textbf{31.6} & \textbf{32.9} & {\color[HTML]{0000FF} \textit{60.3}} & \textbf{50.1} & \textbf{45.1} & \textbf{52.7} & \textbf{225.3 (\textcolor{ao(english)}{+7.1})} \\ \midrule
ST Baselines$\dagger$ &  &  & 82.8 & 78.9 & 81.5 & 60.0 & 83.9 & 26.0 & 55.8 & 34.4 & 22.6 & \textbf{40.4} & 33.5 & 28.4 & 31.4 & 57.5 & 50.0 & 42.8 & 51.0 & 219.7 \\
 &  & \textcolor{lightgray}{\xmark} & \textbf{84.0} & 79.8 & \textbf{82.6} & 45.9 & 85.4 & \textbf{26.3} & 54.2 & 33.3 & 26.8 & 33.3 & 30.5 & 28.5 & 30.4 & 57.5 & 47.8 & 41.0 & 49.7 & 217.0 (\textcolor{purple}{-2.7}) \\
\multirow{-2}{*}{MT Baseline} &  & \textcolor{ao(english)}{\cmark} & 83.5 & \textbf{80.0} & 82.3 & {\color[HTML]{0000FF} \textit{61.7}} & {\color[HTML]{0000FF} \textit{85.6}} & 25.4 & {\color[HTML]{0000FF} \textit{56.5}} & \textbf{35.5} & {\color[HTML]{0000FF} \textit{27.8}} & 34.0 & {\color[HTML]{0000FF} \textit{35.3}} & {\color[HTML]{0000FF} \textit{31.4}} & {\color[HTML]{0000FF} \textit{33.0}} & {\color[HTML]{0000FF} \textit{58.0}} & {\color[HTML]{0000FF} \textit{50.4}} & {\color[HTML]{0000FF} \textit{44.9}} & {\color[HTML]{0000FF} \textit{51.9}} & {\color[HTML]{0000FF} \textit{223.7 (\textcolor{ao(english)}{+4.0})}} \\
\rowcolor{LightCyan} \textbf{\ourmodel} & \multirow{-4}{*}{Swin-T} & \textcolor{ao(english)}{\cmark} & {\color[HTML]{0000FF} \textit{83.8}} & \textbf{80.0} & {\color[HTML]{0000FF} \textit{82.5}} & \textbf{64.5} & \textbf{85.9} & \textbf{26.3} & \textbf{57.5} & {\color[HTML]{0000FF} \textit{35.4}} & \textbf{28.5} & {\color[HTML]{0000FF} \textit{40.2}} & \textbf{35.8} & \textbf{32.2} & \textbf{34.1} & \textbf{60.3} & \textbf{50.5} & \textbf{45.1} & \textbf{52.8} & \textbf{226.9 (\textcolor{ao(english)}{+7.2})} \\
\bottomrule
\end{tabular*}
\vspace{-0.12in}
\end{table*}
\setlength{\tabcolsep}{6pt}

\setlength{\tabcolsep}{8pt}
\begin{table*}[t]
\footnotesize
\centering
\caption{
Comparison of \ourmodelbase with multi-task models and single-task (ST) baselines using ResNet-50 on a subset of \ourbenchmark tasks.
$\dagger$~denotes a separate model is trained for each task.
}
\vspace{-0.15in}
\label{tab:main-mtlmodels}
\begin{tabular}{l|c|ccc|ccccc|cc}
\toprule
 &  & \multicolumn{3}{c|}{Segmentation} & \multicolumn{5}{c|}{Localization} & \multicolumn{2}{c}{Association} \\
 &  & Sem. & Driv. & Lane & Det. & Ins. & Pose & MOT & MOTS & MOT & MOTS \\
\multirow{-3}{*}{Method} & \multirow{-3}{*}{Tasks} & $\text{IoU}^{\texttt{S}}$ & $\text{IoU}^{\texttt{A}}$ & $\text{IoU}^{\texttt{L}}$ & $\text{AP}^{\texttt{D}}$ & $\text{AP}^{\texttt{I}}$ & $\text{AP}^{\texttt{P}}$ & $\text{AP}^{\texttt{T}}$ & $\text{AP}^{\texttt{R}}$ & $\text{AssA}^{\texttt{T}}$ & $\text{AssA}^{\texttt{R}}$ \\ \midrule
ST Baselines$\dagger$ & ST & 59.7 & 83.9 & \textbf{28.4} & 32.3 & 20.2 & 37.0 & 32.9 & 27.2 & 48.8 & 42.4 \\ \midrule
Semantic FPN~\cite{kirillov2019panoptic} & S, A, L & 59.2 & 83.9 & 24.9 & –  & – & – & – & – & – & – \\
Panoptic FPN~\cite{kirillov2019panoptic} & S, I & 58.5 & – & – & – & 19.7 & – & – & – & – & – \\
MaskFormer~\cite{cheng2021per} & S, I & 55.9 & – & – & – & 10.4 & – & – & – & – & – \\
Mask2Former~\cite{cheng2022masked} & S, I & 62.8 & – & – & – & 19.9 & – & – & – & – & – \\
Mask2Former~\cite{cheng2022masked} & S, A, L, I & 59.7 & 84.8 & \textbf{28.4} & – & 17.3 & – & – & – & – & – \\
Mask R-CNN~\cite{he2017mask} & D, P & – & – & – & 32.7 & – & 35.2 & – & – & – & – \\
Mask R-CNN~\cite{he2017mask} & D, I, P & – & – & – & 31.2 & 24.6 & 33.1 & – & – & – & – \\
QDTrack-MOTS~\cite{pang2021quasi} & D, I, T, R & – & – & – & 32.1 & 23.1 & – & 32.9 & 27.2 & 48.8 & 42.4  \\
\midrule
\rowcolor{LightCyan} \textbf{\ourmodel} & \ourbenchmark & \textbf{63.8} & \textbf{85.4} & 27.8 & \textbf{33.4} & \textbf{27.1} & \textbf{39.7} & \textbf{34.7} & \textbf{31.6} & \textbf{50.1} & \textbf{45.1} \\
\bottomrule
\end{tabular}
\vspace{-0.12in}
\end{table*}
\setlength{\tabcolsep}{6pt}

\section{Experiments}
We conduct extensive experiments on \ourbenchmark to evaluate the effectiveness of \ourmodelbase.
We also provide ablation studies and visualizations.

\setlength{\tabcolsep}{4.3pt}
\begin{table*}[]
\scriptsize
\centering
\caption{
Ablation study of network components, including Intra-group (Intra-IB) and Cross-group (Cross-IB) Interaction Blocks with \ourmodelbase using ResNet-50 on \ourbenchmark validation set.
All networks are trained with CPF.
}
\vspace{-0.16in}
\label{tab:ablation-model}
\begin{tabular}{cc|ccc|cccc|cccccc|cccc|c}
\toprule
  &  & \multicolumn{2}{c}{Classification} &  & \multicolumn{3}{c}{Segmentation} &  & \multicolumn{5}{c}{Localization} &  & \multicolumn{3}{c}{Association} &  &  \\
 &  & \multicolumn{2}{c}{Tagging} &  & Sem. & Driv. & Lane &  & Det. & Ins. & Pose & MOT & MOTS &  & Flow & MOT & MOTS &  &  \\
\multirow{-3}{*}{\begin{tabular}{@{}c@{}}Intra-\\IB\end{tabular}} & \multirow{-3}{*}{\begin{tabular}{@{}c@{}}Cross-\\IB\end{tabular}} & $\text{Acc}^{\texttt{Gw}}$ & $\text{Acc}^{\texttt{Gs}}$ & \multirow{-3}{*}{\parbox[t]{1.5mm}{\tiny \rotatebox[origin=c]{90}{VTDA$_{\text{\textcolor{ao(english)}{cls}}}$}}} & $\text{IoU}^{\texttt{S}}$ & $\text{IoU}^{\texttt{A}}$ & $\text{IoU}^{\texttt{L}}$ & \multirow{-3}{*}{\parbox[t]{0mm}{\tiny \rotatebox[origin=c]{90}{VTDA$_{\text{\textcolor{amber}{seg}}}$}}} & $\text{AP}^{\texttt{D}}$ & $\text{AP}^{\texttt{I}}$ & $\text{AP}^{\texttt{P}}$ & $\text{AP}^{\texttt{T}}$ & $\text{AP}^{\texttt{R}}$ & \multirow{-3}{*}{\parbox[t]{0mm}{\tiny \rotatebox[origin=c]{90}{VTDA$_{\text{\textcolor{ceruleanblue}{loc}}}$}}} & $\text{IoU}^{\texttt{F}}$ & $\text{AssA}^{\texttt{T}}$ & $\text{AssA}^{\texttt{R}}$ & \multirow{-3}{*}{\parbox[t]{0mm}{\tiny \rotatebox[origin=c]{90}{VTDA$_{\text{\textcolor{orange}{ass}}}$}}} & \multirow{-3}{*}{VTDA} \\ \midrule
\textcolor{lightgray}{\xmark} & \textcolor{lightgray}{\xmark} & 83.0 & 79.4 & 81.8 & 60.6 & 85.2 & 25.9 & 56.4 & 32.5 & 26.4 & 35.0 & 33.8 & 30.2 & 31.5 & \textbf{60.3} & 49.0 & 43.3 & 51.8 & 221.5 \\
\textcolor{ao(english)}{\cmark} & \textcolor{lightgray}{\xmark} & 82.8 & 79.4 & 81.7 & \textbf{63.9} & 85.2 & 26.0 & 57.0 & \textbf{33.4} & 27.0 & \textbf{39.8} & 34.5 & \textbf{31.7} & 32.8 & 60.2 & 49.3 & \textbf{45.1} & 52.3 & 223.9 (\textcolor{ao(english)}{+2.4}) \\
\rowcolor{LightCyan} \textcolor{ao(english)}{\cmark} & \textcolor{ao(english)}{\cmark} & \textbf{83.2} & \textbf{79.7} & \textbf{82.0} & 63.8 & \textbf{85.4} & \textbf{27.8} & \textbf{57.8} & \textbf{33.4} & \textbf{27.1} & 39.7 & \textbf{34.7} & 31.6 & \textbf{32.9} & \textbf{60.3} & \textbf{50.1} & \textbf{45.1} & \textbf{52.7} & \textbf{225.3 (\textcolor{ao(english)}{+3.8})} \\
\bottomrule
\end{tabular}
\vspace{-0.12in}
\end{table*}
\setlength{\tabcolsep}{6pt}

\setlength{\tabcolsep}{4.5pt}
\begin{table*}[]
\scriptsize
\centering
\caption{
Ablation study of CPF, including curriculum training (C), pseudo-labels (P), and fine-tuning (F) with \ourmodelbase using ResNet-50 on \ourbenchmark validation set.
Highlighted improvements are \underline{underlined.}
}
\vspace{-0.14in}
\label{tab:ablation-cpf}
\begin{tabular}{ccc|ccc|cccc|cccccc|cccc|c}
\toprule
  &  &  & \multicolumn{2}{c}{Classification} &  & \multicolumn{3}{c}{Segmentation} &  & \multicolumn{5}{c}{Localization} &  & \multicolumn{3}{c}{Association} &  &  \\
 &  &  & \multicolumn{2}{c}{Tagging} &  & Sem. & Driv. & Lane &  & Det. & Ins. & Pose & MOT & MOTS &  & Flow & MOT & MOTS &  &  \\
\multirow{-3}{*}{C} & \multirow{-3}{*}{P} & \multirow{-3}{*}{F} & $\text{Acc}^{\texttt{Gw}}$ & $\text{Acc}^{\texttt{Gs}}$ & \multirow{-3}{*}{\parbox[t]{1.5mm}{\tiny \rotatebox[origin=c]{90}{VTDA$_{\text{\textcolor{ao(english)}{cls}}}$}}} & $\text{IoU}^{\texttt{S}}$ & $\text{IoU}^{\texttt{A}}$ & $\text{IoU}^{\texttt{L}}$ & \multirow{-3}{*}{\parbox[t]{0mm}{\tiny \rotatebox[origin=c]{90}{VTDA$_{\text{\textcolor{amber}{seg}}}$}}} & $\text{AP}^{\texttt{D}}$ & $\text{AP}^{\texttt{I}}$ & $\text{AP}^{\texttt{P}}$ & $\text{AP}^{\texttt{T}}$ & $\text{AP}^{\texttt{R}}$ & \multirow{-3}{*}{\parbox[t]{0mm}{\tiny \rotatebox[origin=c]{90}{VTDA$_{\text{\textcolor{ceruleanblue}{loc}}}$}}} & $\text{IoU}^{\texttt{F}}$ & $\text{AssA}^{\texttt{T}}$ & $\text{AssA}^{\texttt{R}}$ & \multirow{-3}{*}{\parbox[t]{0mm}{\tiny \rotatebox[origin=c]{90}{VTDA$_{\text{\textcolor{orange}{ass}}}$}}} & \multirow{-3}{*}{VTDA} \\ \midrule
\textcolor{lightgray}{\xmark} & \textcolor{lightgray}{\xmark} & \textcolor{lightgray}{\xmark} & \textbf{83.6} & 79.1 & \textbf{82.1} & 41.5 & 85.0 & 26.3 & 53.3 & 32.3 & 26.4 & 29.1 & 31.3 & 29.4 & 30.0 & 60.6 & 47.1 & 43.6 & 51.3 & 216.7 \\
\textcolor{ao(english)}{\cmark} & \textcolor{lightgray}{\xmark} & \textcolor{lightgray}{\xmark} & 83.0 & 79.4 & 81.8 & 42.0 & 84.9 & 26.2 & 53.3 & \underline{\textbf{33.9}} & \underline{\textbf{27.3}} & \underline{31.1} & \underline{31.9} & \underline{30.0} & \underline{31.1} & \textbf{60.8} & \underline{47.7} & 43.8 & 51.6 & 217.8 (\textcolor{ao(english)}{+1.1}) \\
\textcolor{ao(english)}{\cmark} & \textcolor{ao(english)}{\cmark} & \textcolor{lightgray}{\xmark} & 83.2 & 79.6 & 82.0 & \underline{63.2} & 85.0 & 26.1 & \underline{56.8} & 33.7 & 27.1 & \underline{39.0} & 31.7 & 30.4 & \underline{31.9} & 60.1 & 48.0 & 44.5 & 51.7 & 222.4 (\textcolor{ao(english)}{+5.7}) \\
\rowcolor{LightCyan} \textcolor{ao(english)}{\cmark} & \textcolor{ao(english)}{\cmark} & \textcolor{ao(english)}{\cmark} & 83.2 & \textbf{79.7} & 82.0 & \underline{\textbf{63.8}} & \underline{\textbf{85.4}} & \underline{\textbf{27.8}} & \underline{\textbf{57.8}} & 33.4 & 27.1 & \underline{\textbf{39.7}} & \underline{\textbf{34.7}} & \underline{\textbf{31.6}} & \underline{\textbf{32.9}} & 60.3 & \underline{\textbf{50.1}} & \underline{\textbf{45.1}} & \underline{\textbf{52.7}} & \textbf{225.3 (\textcolor{ao(english)}{+8.6})} \\
\bottomrule
\end{tabular}
\vspace{-0.14in}
\end{table*}
\setlength{\tabcolsep}{6pt}

\setlength{\tabcolsep}{4pt}
\begin{table}[]
\centering
\footnotesize
\caption{
Ablation study of different loss weight configurations with~\ourmodelbase using ResNet-50 on \ourbenchmark validation set.
}
\vspace{-0.16in}
\label{tab:sup-losses}
\begin{tabular}{@{}c|cccc|c@{}}
\toprule
Loss Weights & VTDA$_{\text{\textcolor{ao(english)}{cls}}}$ & VTDA$_{\text{\textcolor{amber}{seg}}}$ & VTDA$_{\text{\textcolor{ceruleanblue}{loc}}}$ & VTDA$_{\text{\textcolor{orange}{ass}}}$ & VTDA \\
\midrule
Default & 82.0 & 57.8 & 32.9 & 52.7 & \textbf{225.3} \\
$2\lambda_\texttt{G}$ & \textbf{82.2} & 57.6 & 32.4 & 52.5 & 224.7 \\
$2\lambda_\texttt{S},2\lambda_\texttt{A},2\lambda_\texttt{L}$ & 81.9 & \textbf{58.0} & 32.5 & 52.3 & 224.7 \\
$2\lambda_\texttt{D},2\lambda_\texttt{I},2\lambda_\texttt{P}$ & 82.1 & 57.3 & \textbf{33.2} & 52.5 & 225.0 \\
$2\lambda_\texttt{F},2\lambda_\texttt{T}$ & 82.1 & 57.4 & 32.8 & \textbf{53.0} & \textbf{225.3} \\
\bottomrule
\end{tabular}
\vspace{-0.14in}
\end{table}
\setlength{\tabcolsep}{6pt}

\parsection{Implementation Details.}
We use AdamW~\cite{kingma2014adam, loshchilov2017decoupled} with $\beta_1=0.9$, $\beta_2=0.999$, and weight decay of 0.05 as the optimizer.
We initialize all models with ImageNet pre-trained weights~\cite{imagenet}.
All models are trained for 12 epochs with a batch size of 16 and full crop size of $720 \times 1280$.
We use a learning rate of $0.0001$, decreasing by a factor of 10 at epochs 8 and 11.
For augmentation, we use multi-scale training and flipping.
We use the same data augmentations and learning rate for every task to reduce efforts needed for hyperparameter tuning.

\subsection{Main Results} \label{sec:main-res}
We compare multi-task performance and computational efficiency of \ourmodelbase against single-task and multi-task baselines as well as other multi-task models on \ourbenchmark.

\parsection{Comparison to Baselines.}
For fair comparison, we first compare \ourmodelbase to two baselines: single-task and multi-task.
Single-task baselines use the same architecture described in \sectionautorefname~\ref{section-model} for each task without the extra components used for other tasks and without interaction blocks.
Each single-task baseline is trained solely on the corresponding task data without using other annotations, except for MOT/MOTS that uses detection/instance segmentation data during training.
We do not fix any other component besides the architecture and data,
and we train each with task-specific augmentations and learning rate schedules, optimized for single-task performance.
The multi-task baseline also uses the same architecture without interaction blocks, but it is optimized on all ten tasks jointly and uses all the \ourbenchmark data.
We provide complete training details of every baseline in \sectionautorefname~\ref{sec:sup-tdet}.

\begin{figure}[t]
\centering
   \begin{subfigure}{0.28\linewidth} \centering
     \includegraphics[trim={0 0 3.5cm 0},clip,width=\linewidth]{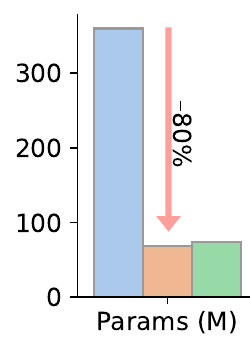}
   \end{subfigure}
   \begin{subfigure}{0.675\linewidth} \centering
     \includegraphics[trim={0 0 0.3cm 0},clip,width=\linewidth]{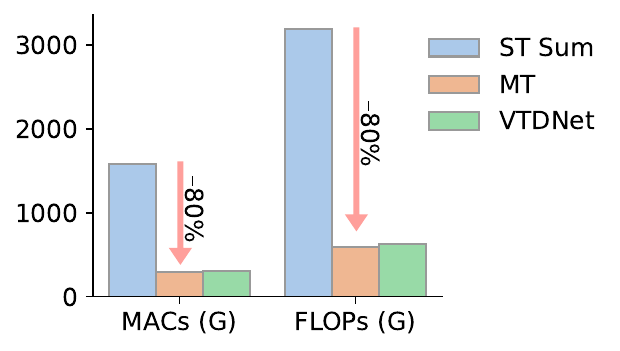}
   \end{subfigure}
\vspace{-0.18in}
\caption{
Comparison of resource usage during inference between \ourmodelbase, single-task (ST), and multi-task (MT) baselines using ResNet-50 as the base network.
Compared to ST baselines, \ourmodelbase uses only one-fifth of the resources.
}
\label{fig:params}
\vspace{-0.16in}
\end{figure}

We conduct experiments on two different base networks: ResNet-50~\cite{he2016deep} and Swin Transformer (Swin-T)~\cite{liu2021swin}.
The results are shown in \tableautorefname~\ref{tab:main-results-backbones}.
First, we find naive joint training with the multi-task baseline to not bring improvements to overall multi-task performance over the single-task baselines and severely hurts the accuracy of some tasks due to label deficiency (\eg, pose estimation and semantic segmentation), task interference (\eg, lane detection), and under-training (\eg, MOT).
This shows that a sophisticated training strategy is necessary to overcome optimization challenges in \ourbenchmark.
When optimizing the multi-task baseline with our \ourlearn training protocol, significant performance gains can be achieved across the board, obtaining better scores than the single-task baselines on a majority of tasks and an improvement of over 3 points in \ourmetric.
Furthermore,
\ourmodelbase is able to achieve additional improvements in performance over the baselines, obtaining the best scores on most tasks and an increase of over 7 points in \ourmetric across both base networks.

\parsection{Comparison to Multi-Task Models.}
We compare \ourmodelbase against various other MTL models trained on a subset of the VTD tasks in \tableautorefname~\ref{tab:main-mtlmodels}.
We train these models with set-level batch sampling and task-specific data augmentations and schedules.
While leveraging additional data from other tasks in a multi-task learning setting can boost per-task performance of a few tasks, performance on certain tasks (lane detection and pose estimation) greatly suffers due to task interference and label deficiency.
In particular, we find Mask2Former~\cite{cheng2022masked} to perform well on semantic segmentation, but its instance segmentation performance is worse as it cannot take advantage of the abundant bounding box annotations – adding a detection decoder results in extremely poor detection accuracy of 17.8 AP$^\texttt{D}$.
We further extend Mask2Former to handle other segmentation tasks and find it performs competitively, though performance on original tasks are degraded.
In comparison, \ourmodelbase can achieve further performance gains across all tasks while alleviating the performance loss, demonstrating the benefits of unifying the \ourbenchmark tasks.

\parsection{Resource Usage.}
We compare the resource usage during inference of \ourmodelbase, single-task, and multi-task baselines with ResNet-50 base network in \figureautorefname~\ref{fig:params}.
Since each single-task baseline uses a separate feature extractor, the computation accumulates with the increasing number of tasks.
Comparatively, the multi-task baseline and \ourmodelbase use around 80\% fewer model parameters and operations due to sharing the feature extractor among all tasks.
Additionally, since \ourmodel replaces independent decoding layers in each task decoder with shared feature interaction blocks, \ourmodel achieves significantly better performance with negligible computational overhead compared to the multi-task baseline.

\subsection{Ablation Study and Analysis}
We conduct a variety of ablation studies to evaluate different aspects of our network and our training protocol.

\parsection{Network Components.}
We compare the effect of our feature interaction blocks, Intra-IB and Cross-IB, on \ourmodelbase in \tableautorefname~\ref{tab:ablation-model}.
We train all networks with CPF for a fair comparison.
Adding Intra-IB to model feature interactions between tasks in the same groups leads to significant improvements across segmentation and localization tasks, which results in an overall increase of 2.5 points in \ourmetric.
In particular, performance on label-deficient tasks, semantic segmentation and pose estimation, is improved by over 3 points.
Using Cross-IB further improves performance on segmentation tasks by 0.8 points on average, which leads to an additional 1.4 points increase in \ourmetric.
This demonstrates that additional feature sharing within and between task groups can both largely benefit heterogeneous multi-task performance.

\parsection{Training Protocol.}
We evaluate how components of our \ourlearn protocol affect the final \ourmodelbase performance on \ourbenchmark in \tableautorefname~\ref{tab:ablation-cpf}.
Curriculum training significantly improves localization performance by pre-training the network before joint optimization.
Using pose estimation and semantic segmentation pseudo-labels can completely resolve the label deficiency issue and result in much better performance on those tasks.
Fine-tuning can further improve scores across the board by optimizing on task-specific data, especially for object tracking and segmentation tasks.
By utilizing \ourlearn, we can handle the optimization challenges of \ourbenchmark and bring out the true benefits of multi-task learning.

\begin{figure*}[t]
\centering
\includegraphics[width=.99\linewidth]{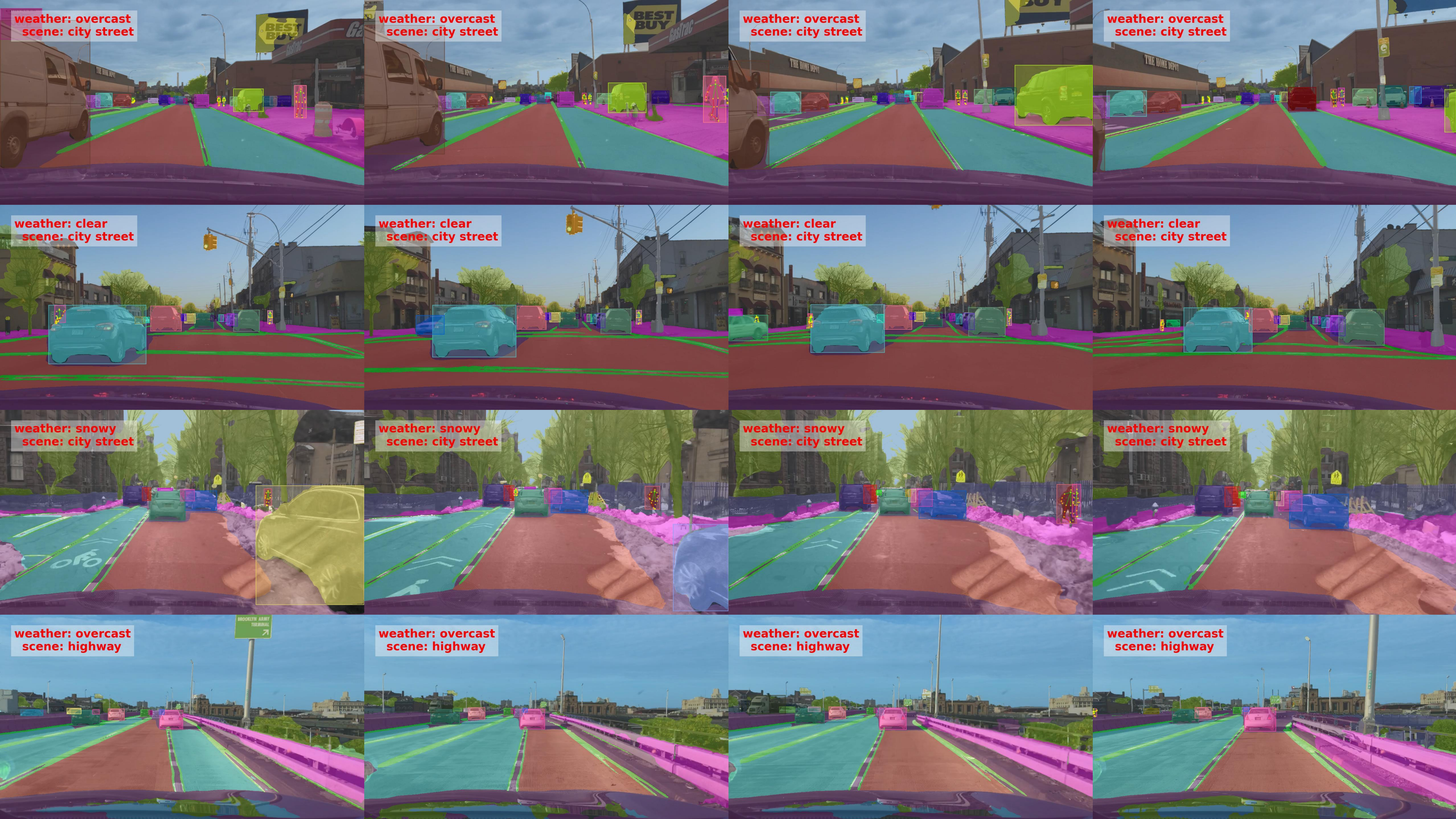}
\vspace{-0.12in}
\caption{
Visualization of \ourmodel predictions on all tasks (excluding flow).
Best viewed in color and zoomed in.
}
\label{fig:vtd-vis}
\vspace{-0.14in}
\end{figure*}

\parsection{Loss Weights and Metric.}
We investigate how \ourmodelbase and \ourmetric behaves with various loss weight configurations in \tableautorefname~\ref{tab:sup-losses}.
Increasing or decreasing task loss weights results in the corresponding task group performance to increase or decrease, showing that one can modify the loss weights depending on the application to prioritize performance on certain tasks.
\ourmetric can clearly demonstrate the improvements and decreases in performance of different aspects of the network.
Furthermore, \ourmetric remains relatively stable across different configurations.

\parsection{Visualizations.}
We show qualitative results of \ourmodel on VTD validation set for several video sequences in \figureautorefname~\ref{fig:vtd-vis}.
The predictions of each task (excluding flow) are overlaid on top of each other in each frame.
The color of each object indicate the predicted instance identity.
For drivable area segmentation, red areas on the road indicate drivable regions, and blue areas indicate alternatively drivable areas.
The green lines represent predicted lanes on the road.
Across all sequences, \ourmodel can produce high-quality predictions that are consistent across all ten tasks with a single forward pass on each image.

\section{Discussion and Conclusions}
In this work, we present our new \ourbenchmarkfull challenge to study heterogeneous multi-task learning for autonomous driving.
\ourbenchmark includes ten representative tasks on images and videos, allowing for the exploration of a unified representation for 2D vision.
Our heterogeneous multi-task model \ourmodelbase, equipped with feature interaction blocks and our CPF training protocol, significantly enhances the performance of single-task models while being much more efficient.
We hope the \ourbenchmark challenge can spark interest in this important area of research.


{\small
\bibliographystyle{ieee_fullname}
\bibliography{main}

\begin{thebibliography}{10}\itemsep=-1pt

\bibitem{ba2016layer}
Jimmy~Lei Ba, Jamie~Ryan Kiros, and Geoffrey~E Hinton.
\newblock Layer normalization.
\newblock {\em arXiv preprint arXiv:1607.06450}, 2016.

\bibitem{nuscenes2019}
Holger Caesar, Varun Bankiti, Alex~H. Lang, Sourabh Vora, Venice~Erin Liong, Qiang Xu, Anush Krishnan, Yu Pan, Giancarlo Baldan, and Oscar Beijbom.
\newblock nuscenes: A multimodal dataset for autonomous driving.
\newblock {\em arXiv preprint arXiv:1903.11027}, 2019.

\bibitem{cai2018cascade}
Zhaowei Cai and Nuno Vasconcelos.
\newblock Cascade r-cnn: Delving into high quality object detection.
\newblock In {\em CVPR}, pages 6154--6162, 2018.

\bibitem{caruana1997multitask}
Rich Caruana.
\newblock Multitask learning.
\newblock {\em Machine learning}, 1997.

\bibitem{chen2019hybrid}
Kai Chen, Jiangmiao Pang, Jiaqi Wang, Yu Xiong, Xiaoxiao Li, Shuyang Sun, Wansen Feng, Ziwei Liu, Jianping Shi, Wanli Ouyang, et~al.
\newblock Hybrid task cascade for instance segmentation.
\newblock In {\em Proceedings of the IEEE/CVF conference on computer vision and pattern recognition}, pages 4974--4983, 2019.

\bibitem{chen2018encoder}
Liang-Chieh Chen, Yukun Zhu, George Papandreou, Florian Schroff, and Hartwig Adam.
\newblock Encoder-decoder with atrous separable convolution for semantic image segmentation.
\newblock In {\em Proceedings of the European conference on computer vision (ECCV)}, pages 801--818, 2018.

\bibitem{chen2018gradnorm}
Zhao Chen, Vijay Badrinarayanan, Chen-Yu Lee, and Andrew Rabinovich.
\newblock Gradnorm: Gradient normalization for adaptive loss balancing in deep multitask networks.
\newblock In {\em ICML}, pages 794--803. PMLR, 2018.

\bibitem{cheng2022masked}
Bowen Cheng, Ishan Misra, Alexander~G Schwing, Alexander Kirillov, and Rohit Girdhar.
\newblock Masked-attention mask transformer for universal image segmentation.
\newblock In {\em CVPR}, pages 1290--1299, 2022.

\bibitem{cheng2021per}
Bowen Cheng, Alex Schwing, and Alexander Kirillov.
\newblock Per-pixel classification is not all you need for semantic segmentation.
\newblock {\em NeurIPS}, 34, 2021.

\bibitem{Cordts2016Cityscapes}
Marius Cordts, Mohamed Omran, Sebastian Ramos, Timo Rehfeld, Markus Enzweiler, Rodrigo Benenson, Uwe Franke, Stefan Roth, and Bernt Schiele.
\newblock The cityscapes dataset for semantic urban scene understanding.
\newblock In {\em CVPR}, 2016.

\bibitem{dendorfer2020mot20}
Patrick Dendorfer, Hamid Rezatofighi, Anton Milan, Javen Shi, Daniel Cremers, Ian Reid, Stefan Roth, Konrad Schindler, and Laura Leal-Taix{\'e}.
\newblock Mot20: A benchmark for multi object tracking in crowded scenes.
\newblock {\em arXiv preprint arXiv:2003.09003}, 2020.

\bibitem{imagenet}
Jia Deng, Wei Dong, Richard Socher, Li-Jia Li, Kai Li, and Li Fei-Fei.
\newblock Imagenet: A large-scale hierarchical image database.
\newblock In {\em CVPR}, pages 248--255, 2009.

\bibitem{fischer2022qdtrack}
Tobias Fischer, Thomas~E Huang, Jiangmiao Pang, Linlu Qiu, Haofeng Chen, Trevor Darrell, and Fisher Yu.
\newblock Qdtrack: quasi-dense similarity learning for appearance-only multiple object tracking.
\newblock {\em IEEE Transactions on Pattern Analysis and Machine Intelligence}, 2023.

\bibitem{Geiger2012CVPR}
Andreas Geiger, Philip Lenz, and Raquel Urtasun.
\newblock Are we ready for autonomous driving? the kitti vision benchmark suite.
\newblock In {\em CVPR}, 2012.

\bibitem{ghiasi2021multi}
Golnaz Ghiasi, Barret Zoph, Ekin~D Cubuk, Quoc~V Le, and Tsung-Yi Lin.
\newblock Multi-task self-training for learning general representations.
\newblock In {\em Proceedings of the IEEE/CVF International Conference on Computer Vision}, pages 8856--8865, 2021.

\bibitem{Goel_2021_WACV}
Kratarth Goel, Praveen Srinivasan, Sarah Tariq, and James Philbin.
\newblock Quadronet: Multi-task learning for real-time semantic depth aware instance segmentation.
\newblock In {\em WACV}, pages 315--324, January 2021.

\bibitem{Guo_2018_ECCV}
Michelle Guo, Albert Haque, De-An Huang, Serena Yeung, and Li Fei-Fei.
\newblock Dynamic task prioritization for multitask learning.
\newblock In {\em ECCV}, September 2018.

\bibitem{he2017mask}
Kaiming He, Georgia Gkioxari, Piotr Doll{\'a}r, and Ross Girshick.
\newblock Mask r-cnn.
\newblock In {\em ICCV}, pages 2961--2969, 2017.

\bibitem{he2016deep}
Kaiming He, Xiangyu Zhang, Shaoqing Ren, and Jian Sun.
\newblock Deep residual learning for image recognition.
\newblock In {\em CVPR}, pages 770--778, 2016.

\bibitem{kanakis2020reparameterizing}
Menelaos Kanakis, David Bruggemann, Suman Saha, Stamatios Georgoulis, Anton Obukhov, and Luc~Van Gool.
\newblock Reparameterizing convolutions for incremental multi-task learning without task interference.
\newblock In {\em ECCV}. Springer, 2020.

\bibitem{kanakis2022composite}
Menelaos Kanakis, Thomas~E Huang, David Bruggemann, Fisher Yu, and Luc Van~Gool.
\newblock Composite learning for robust and effective dense predictions.
\newblock {\em arXiv preprint arXiv:2210.07239}, 2022.

\bibitem{kanakis2023composite}
Menelaos Kanakis, Thomas~E Huang, David Br{\"u}ggemann, Fisher Yu, and Luc Van~Gool.
\newblock Composite learning for robust and effective dense predictions.
\newblock In {\em Proceedings of the IEEE/CVF Winter Conference on Applications of Computer Vision}, pages 2299--2308, 2023.

\bibitem{transfiner}
Lei Ke, Martin Danelljan, Xia Li, Yu-Wing Tai, Chi-Keung Tang, and Fisher Yu.
\newblock Mask transfiner for high-quality instance segmentation.
\newblock In {\em CVPR}, 2022.

\bibitem{ke2022video}
Lei Ke, Henghui Ding, Martin Danelljan, Yu-Wing Tai, Chi-Keung Tang, and Fisher Yu.
\newblock Video mask transfiner for high-quality video instance segmentation.
\newblock In {\em Computer Vision--ECCV 2022: 17th European Conference, Tel Aviv, Israel, October 23--27, 2022, Proceedings, Part XXVIII}, pages 731--747. Springer, 2022.

\bibitem{ke2021prototypical}
Lei Ke, Xia Li, Martin Danelljan, Yu-Wing Tai, Chi-Keung Tang, and Fisher Yu.
\newblock Prototypical cross-attention networks for multiple object tracking and segmentation.
\newblock {\em NeurIPS}, 34, 2021.

\bibitem{kendall2018multi}
Alex Kendall, Yarin Gal, and Roberto Cipolla.
\newblock Multi-task learning using uncertainty to weigh losses for scene geometry and semantics.
\newblock In {\em CVPR}, 2018.

\bibitem{kingma2014adam}
Diederik~P Kingma and Jimmy Ba.
\newblock Adam: A method for stochastic optimization.
\newblock {\em arXiv preprint arXiv:1412.6980}, 2014.

\bibitem{kirillov2019panoptic}
Alexander Kirillov, Ross Girshick, Kaiming He, and Piotr Doll{\'a}r.
\newblock Panoptic feature pyramid networks.
\newblock In {\em CVPR}, 2019.

\bibitem{kokkinos2017ubernet}
Iasonas Kokkinos.
\newblock Ubernet: Training a universal convolutional neural network for low-, mid-, and high-level vision using diverse datasets and limited memory.
\newblock In {\em CVPR}, pages 6129--6138, 2017.

\bibitem{teter}
Siyuan Li, Martin Danelljan, Henghui Ding, Thomas~E. Huang, and Fisher Yu.
\newblock Tracking every thing in the wild.
\newblock In {\em ECCV}, 2022.

\bibitem{liang2022effective}
Xiwen Liang, Yangxin Wu, Jianhua Han, Hang Xu, Chunjing Xu, and Xiaodan Liang.
\newblock Effective adaptation in multi-task co-training for unified autonomous driving.
\newblock {\em neurips}, 2022.

\bibitem{likhosherstov2021polyvit}
Valerii Likhosherstov, Anurag Arnab, Krzysztof Choromanski, Mario Lucic, Yi Tay, Adrian Weller, and Mostafa Dehghani.
\newblock Polyvit: Co-training vision transformers on images, videos and audio.
\newblock {\em arXiv preprint arXiv:2111.12993}, 2021.

\bibitem{lin2017feature}
Tsung-Yi Lin, Piotr Doll{\'a}r, Ross Girshick, Kaiming He, Bharath Hariharan, and Serge Belongie.
\newblock Feature pyramid networks for object detection.
\newblock In {\em CVPR}, 2017.

\bibitem{liu2021towards}
Liyang Liu, Yi Li, Zhanghui Kuang, Jing{-}Hao Xue, Yimin Chen, Wenming Yang, Qingmin Liao, and Wayne Zhang.
\newblock Towards impartial multi-task learning.
\newblock In {\em 9th International Conference on Learning Representations, {ICLR} 2021, Virtual Event, Austria, May 3-7, 2021}. OpenReview.net, 2021.

\bibitem{liu2021swin}
Ze Liu, Yutong Lin, Yue Cao, Han Hu, Yixuan Wei, Zheng Zhang, Stephen Lin, and Baining Guo.
\newblock Swin transformer: Hierarchical vision transformer using shifted windows.
\newblock In {\em ICCV}, pages 10012--10022, 2021.

\bibitem{liu2022convnet}
Zhuang Liu, Hanzi Mao, Chao-Yuan Wu, Christoph Feichtenhofer, Trevor Darrell, and Saining Xie.
\newblock A convnet for the 2020s.
\newblock {\em CVPR}, 2022.

\bibitem{loshchilov2017decoupled}
Ilya Loshchilov and Frank Hutter.
\newblock Decoupled weight decay regularization.
\newblock {\em iclr}, 2019.

\bibitem{luiten2021hota}
Jonathon Luiten, Aljosa Osep, Patrick Dendorfer, Philip Torr, Andreas Geiger, Laura Leal-Taix{\'e}, and Bastian Leibe.
\newblock Hota: A higher order metric for evaluating multi-object tracking.
\newblock {\em IJCV}, 129(2):548--578, 2021.

\bibitem{maninis2019attentive}
Kevis-Kokitsi Maninis, Ilija Radosavovic, and Iasonas Kokkinos.
\newblock Attentive single-tasking of multiple tasks.
\newblock In {\em CVPR}, pages 1851--1860, 2019.

\bibitem{meister2018unflow}
Simon Meister, Junhwa Hur, and Stefan Roth.
\newblock Unflow: Unsupervised learning of optical flow with a bidirectional census loss.
\newblock In {\em AAAI}, 2018.

\bibitem{milan2016mot16}
Anton Milan, Laura Leal-Taix{\'e}, Ian Reid, Stefan Roth, and Konrad Schindler.
\newblock Mot16: A benchmark for multi-object tracking.
\newblock {\em arXiv preprint arXiv:1603.00831}, 2016.

\bibitem{mottaghi_cvpr14}
Roozbeh Mottaghi, Xianjie Chen, Xiaobai Liu, Nam-Gyu Cho, Seong-Whan Lee, Sanja Fidler, Raquel Urtasun, and Alan Yuille.
\newblock The role of context for object detection and semantic segmentation in the wild.
\newblock In {\em CVPR}, 2014.

\bibitem{Silberman:ECCV12}
Pushmeet~Kohli Nathan~Silberman, Derek~Hoiem and Rob Fergus.
\newblock Indoor segmentation and support inference from rgbd images.
\newblock In {\em ECCV}, 2012.

\bibitem{pang2021quasi}
Jiangmiao Pang, Linlu Qiu, Xia Li, Haofeng Chen, Qi Li, Trevor Darrell, and Fisher Yu.
\newblock Quasi-dense similarity learning for multiple object tracking.
\newblock In {\em CVPR}, 2021.

\bibitem{rebuffi2017learning}
Sylvestre-Alvise Rebuffi, Hakan Bilen, and Andrea Vedaldi.
\newblock Learning multiple visual domains with residual adapters.
\newblock {\em NeurIPS}, 30, 2017.

\bibitem{ren2015faster}
Shaoqing Ren, Kaiming He, Ross Girshick, and Jian Sun.
\newblock Faster r-cnn: Towards real-time object detection with region proposal networks.
\newblock {\em NeurIPS}, 28, 2015.

\bibitem{ristani2016performance}
Ergys Ristani, Francesco Solera, Roger Zou, Rita Cucchiara, and Carlo Tomasi.
\newblock Performance measures and a data set for multi-target, multi-camera tracking.
\newblock In {\em ECCV}, pages 17--35. Springer, 2016.

\bibitem{SGD}
Herbert Robbins and Sutton Monro.
\newblock {A Stochastic Approximation Method}.
\newblock {\em The Annals of Mathematical Statistics}, 22(3):400 -- 407, 1951.

\bibitem{ruder2017overview}
Sebastian Ruder.
\newblock An overview of multi-task learning in deep neural networks.
\newblock {\em arXiv preprint arXiv:1706.05098}, 2017.

\bibitem{saleh2018effective}
Fatemeh~Sadat Saleh, Mohammad~Sadegh Aliakbarian, Mathieu Salzmann, Lars Petersson, and Jose~M Alvarez.
\newblock Effective use of synthetic data for urban scene semantic segmentation.
\newblock In {\em Proceedings of the European Conference on Computer Vision (ECCV)}, pages 84--100, 2018.

\bibitem{Schon_2021_ICCV}
Markus Sch\"on, Michael Buchholz, and Klaus Dietmayer.
\newblock Mgnet: Monocular geometric scene understanding for autonomous driving.
\newblock In {\em ICCV}, 2021.

\bibitem{sener2018multi}
Ozan Sener and Vladlen Koltun.
\newblock Multi-task learning as multi-objective optimization.
\newblock {\em NeurIPS}, 31, 2018.

\bibitem{stief2007clear}
Rainer Stiefelhagen, Keni Bernardin, Rachel Bowers, John Garofolo, Djamel Mostefa, and Padmanabhan Soundararajan.
\newblock The clear 2006 evaluation.
\newblock In {\em Multimodal Technologies for Perception of Humans}, pages 1--44, Berlin, Heidelberg, 2007. Springer Berlin Heidelberg.

\bibitem{sun2018pwc}
Deqing Sun, Xiaodong Yang, Ming-Yu Liu, and Jan Kautz.
\newblock Pwc-net: Cnns for optical flow using pyramid, warping, and cost volume.
\newblock In {\em CVPR}, 2018.

\bibitem{sun2020waymo}
Pei Sun, Henrik Kretzschmar, Xerxes Dotiwalla, Aurelien Chouard, Vijaysai Patnaik, Paul Tsui, James Guo, Yin Zhou, Yuning Chai, Benjamin Caine, Vijay Vasudevan, Wei Han, Jiquan Ngiam, Hang Zhao, Aleksei Timofeev, Scott Ettinger, Maxim Krivokon, Amy Gao, Aditya Joshi, Yu Zhang, Jonathon Shlens, Zhifeng Chen, and Dragomir Anguelov.
\newblock Scalability in perception for autonomous driving: Waymo open dataset.
\newblock In {\em Proceedings of the IEEE/CVF Conference on Computer Vision and Pattern Recognition (CVPR)}, June 2020.

\bibitem{sun2022shift}
Tao Sun, Mattia Segu, Janis Postels, Yuxuan Wang, Luc Van~Gool, Bernt Schiele, Federico Tombari, and Fisher Yu.
\newblock {SHIFT:} a synthetic driving dataset for continuous multi-task domain adaptation.
\newblock In {\em CVPR}, 2022.

\bibitem{sun2020adashare}
Ximeng Sun, Rameswar Panda, Rogerio Feris, and Kate Saenko.
\newblock Adashare: Learning what to share for efficient deep multi-task learning.
\newblock {\em NeurIPS}, 2020.

\bibitem{tomasi1998bilateral}
C. Tomasi and R. Manduchi.
\newblock Bilateral filtering for gray and color images.
\newblock In {\em Sixth International Conference on Computer Vision (IEEE Cat. No.98CH36271)}, pages 839--846, 1998.

\bibitem{hsuan2016objectflow}
Yi-Hsuan Tsai, Ming-Hsuan Yang, and Michael~J. Black.
\newblock Video segmentation via object flow.
\newblock In {\em CVPR}, pages 3899--3908, 2016.

\bibitem{vandenhende2020mti}
Simon Vandenhende, Stamatios Georgoulis, and Luc~Van Gool.
\newblock Mti-net: Multi-scale task interaction networks for multi-task learning.
\newblock In {\em ECCV}, 2020.

\bibitem{vaswani2017attention}
Ashish Vaswani, Noam Shazeer, Niki Parmar, Jakob Uszkoreit, Llion Jones, Aidan~N Gomez, {\L}ukasz Kaiser, and Illia Polosukhin.
\newblock Attention is all you need.
\newblock {\em NeurIPS}, 30, 2017.

\bibitem{wang2018occlusion}
Yang Wang, Yi Yang, Zhenheng Yang, Liang Zhao, Peng Wang, and Wei Xu.
\newblock Occlusion aware unsupervised learning of optical flow.
\newblock In {\em CVPR}, pages 4884--4893, 2018.

\bibitem{wu2021yolop}
Dong Wu, Manwen Liao, Weitian Zhang, and Xinggang Wang.
\newblock Yolop: You only look once for panoptic driving perception.
\newblock {\em arXiv preprint arXiv:2108.11250}, 2021.

\bibitem{wu2018group}
Yuxin Wu and Kaiming He.
\newblock Group normalization.
\newblock In {\em ECCV}, pages 3--19, 2018.

\bibitem{Xiao_2018_ECCV}
Bin Xiao, Haiping Wu, and Yichen Wei.
\newblock Simple baselines for human pose estimation and tracking.
\newblock In {\em ECCV}, September 2018.

\bibitem{xiao2018unified}
Tete Xiao, Yingcheng Liu, Bolei Zhou, Yuning Jiang, and Jian Sun.
\newblock Unified perceptual parsing for scene understanding.
\newblock In {\em ECCV}, pages 418--434, 2018.

\bibitem{xie2020self}
Qizhe Xie, Minh-Thang Luong, Eduard Hovy, and Quoc~V Le.
\newblock Self-training with noisy student improves imagenet classification.
\newblock In {\em Proceedings of the IEEE/CVF conference on computer vision and pattern recognition}, pages 10687--10698, 2020.

\bibitem{Xu_2018_CVPR}
Dan Xu, Wanli Ouyang, Xiaogang Wang, and Nicu Sebe.
\newblock Pad-net: Multi-tasks guided prediction-and-distillation network for simultaneous depth estimation and scene parsing.
\newblock In {\em CVPR}, June 2018.

\bibitem{xu2022mtformer}
Xiaogang Xu, Hengshuang Zhao, Vibhav Vineet, Ser-Nam Lim, and Antonio Torralba.
\newblock Mtformer: Multi-task learning via transformer and cross-task reasoning.
\newblock In {\em Computer Vision – ECCV 2022: 17th European Conference, Tel Aviv, Israel, October 23–27, 2022, Proceedings, Part XXVII}, page 304–321, Berlin, Heidelberg, 2022. Springer-Verlag.

\bibitem{unicorn}
Bin Yan, Yi Jiang, Peize Sun, Dong Wang, Zehuan Yuan, Ping Luo, and Huchuan Lu.
\newblock Towards grand unification of object tracking.
\newblock In {\em ECCV}, 2022.

\bibitem{yang2023dense}
Yung-Hsu Yang, Thomas~E Huang, Min Sun, Samuel~Rota Bul{\`o}, Peter Kontschieder, and Fisher Yu.
\newblock Dense prediction with attentive feature aggregation.
\newblock In {\em Proceedings of the IEEE/CVF Winter Conference on Applications of Computer Vision}, pages 97--106, 2023.

\bibitem{ye2022inverted}
Hanrong Ye and Dan Xu.
\newblock Inverted pyramid multi-task transformer for dense scene understanding.
\newblock In {\em Computer Vision--ECCV 2022: 17th European Conference, Tel Aviv, Israel, October 23--27, 2022, Proceedings, Part XXVII}, pages 514--530. Springer, 2022.

\bibitem{ye2023taskprompter}
Hanrong Ye and Dan Xu.
\newblock Taskprompter: Spatial-channel multi-task prompting for dense scene understanding.
\newblock In {\em The Eleventh International Conference on Learning Representations}, 2023.

\bibitem{yin2020disentangled}
Minghao Yin, Zhuliang Yao, Yue Cao, Xiu Li, Zheng Zhang, Stephen Lin, and Han Hu.
\newblock Disentangled non-local neural networks, 2020.

\bibitem{yu2020bdd100k}
Fisher Yu, Haofeng Chen, Xin Wang, Wenqi Xian, Yingying Chen, Fangchen Liu, Vashisht Madhavan, and Trevor Darrell.
\newblock Bdd100k: A diverse driving dataset for heterogeneous multitask learning.
\newblock In {\em CVPR}, pages 2636--2645, 2020.

\bibitem{yu2020gradient}
Tianhe Yu, Saurabh Kumar, Abhishek Gupta, Sergey Levine, Karol Hausman, and Chelsea Finn.
\newblock Gradient surgery for multi-task learning.
\newblock In H. Larochelle, M. Ranzato, R. Hadsell, M.~F. Balcan, and H. Lin, editors, {\em NeurIPS}, volume~33, pages 5824--5836. Curran Associates, Inc., 2020.

\bibitem{yuan2021florence}
Lu Yuan, Dongdong Chen, Yi-Ling Chen, Noel Codella, Xiyang Dai, Jianfeng Gao, Houdong Hu, Xuedong Huang, Boxin Li, Chunyuan Li, et~al.
\newblock Florence: A new foundation model for computer vision.
\newblock {\em arXiv preprint arXiv:2111.11432}, 2021.

\bibitem{zamir2018taskonomy}
Amir~R Zamir, Alexander Sax, William Shen, Leonidas~J Guibas, Jitendra Malik, and Silvio Savarese.
\newblock Taskonomy: Disentangling task transfer learning.
\newblock In {\em CVPR}, 2018.

\bibitem{zhang2023demt}
Lefei Zhang et~al.
\newblock Demt: Deformable mixer transformer for multi-task learning of dense prediction.
\newblock {\em arXiv preprint arXiv:2301.03461}, 2023.

\bibitem{zhang2019pattern}
Zhenyu Zhang, Zhen Cui, Chunyan Xu, Yan Yan, Nicu Sebe, and Jian Yang.
\newblock Pattern-affinitive propagation across depth, surface normal and semantic segmentation.
\newblock In {\em CVPR}, pages 4106--4115, 2019.

\bibitem{zhu2019deformable}
Xizhou Zhu, Han Hu, Stephen Lin, and Jifeng Dai.
\newblock Deformable convnets v2: More deformable, better results.
\newblock In {\em CVPR}, pages 9308--9316, 2019.

\bibitem{zhu2020deformable}
Xizhou Zhu, Weijie Su, Lewei Lu, Bin Li, Xiaogang Wang, and Jifeng Dai.
\newblock Deformable detr: Deformable transformers for end-to-end object detection.
\newblock {\em arXiv preprint arXiv:2010.04159}, 2020.

\end{thebibliography}
}

\clearpage
\appendix

\section*{Appendix}

The appendix is organized as follows:
\begin{itemize}
\itemsep-0.3em
    \item Section~\ref{sec:sup-basenetwork}: Additional base networks
    \item Section~\ref{sec:sup-singletask}: Full single-task comparison
    \item Section~\ref{sec:sup-kitti}: Experiments on KITTI dataset
    \item Section~\ref{sec:sup-abs}: Additional ablation studies
    \item Section~\ref{sec:sup-res}: Full resource usage comparison
    \item Section~\ref{sec:sup-benchmark}: \ourbenchmark challenge details
    \item Section~\ref{sec:sup-hdet}: \ourmodel details
    \item Section~\ref{sec:sup-cpf}: \ourlearn training protocol details
    \item Section~\ref{sec:sup-tdet}: Training details
    \item Section~\ref{sec:sup-vis}: Additional visualizations
\end{itemize}

\definecolor{LightCyan}{rgb}{0.88,1,1}
\setlength{\tabcolsep}{3.2pt}
\begin{table*}[]
\scriptsize
\centering
\caption{
Comparison of \ourmodelbase against single-task (ST) baselines using additional base networks on VTD validation set.
$\dagger$~denotes a separate model is trained for each task.
\ourmodelbase outperforms ST baselines on most tasks across all base networks and achieves significantly better \ourmetric.
}
\label{tab:sup-backbones}
\vspace{-0.15in}
\begin{tabular}{lc|ccc|cccc|cccccc|cccc|c}
\toprule
 &  & \multicolumn{2}{c}{Classification} &  & \multicolumn{3}{c}{Segmentation} &  & \multicolumn{5}{c}{Localization} &  & \multicolumn{3}{c}{Association} &  &  \\
 & & \multicolumn{2}{c}{Tagging} &  & Sem. & Driv. & Lane &  & Det. & Ins. & Pose & MOT & MOTS &  & Flow & MOT & MOTS &  &  \\
\multirow{-3}{*}{Method} & \multirow{-3}{*}{\begin{tabular}{@{}c@{}}Base\\Network\end{tabular}} & $\text{Acc}^{\texttt{Gw}}$ & $\text{Acc}^{\texttt{Gs}}$ & \multirow{-3}{*}{\parbox[t]{1.5mm}{\tiny \rotatebox[origin=c]{90}{VTDA$_{\text{\textcolor{ao(english)}{cls}}}$}}} & $\text{IoU}^{\texttt{S}}$ & $\text{IoU}^{\texttt{A}}$ & $\text{IoU}^{\texttt{L}}$ & \multirow{-3}{*}{\parbox[t]{0mm}{\tiny \rotatebox[origin=c]{90}{VTDA$_{\text{\textcolor{amber}{seg}}}$}}} & $\text{AP}^{\texttt{D}}$ & $\text{AP}^{\texttt{I}}$ & $\text{AP}^{\texttt{P}}$ & $\text{AP}^{\texttt{T}}$ & $\text{AP}^{\texttt{R}}$ & \multirow{-3}{*}{\parbox[t]{0mm}{\tiny \rotatebox[origin=c]{90}{VTDA$_{\text{\textcolor{ceruleanblue}{loc}}}$}}} & $\text{IoU}^{\texttt{F}}$ & $\text{AssA}^{\texttt{T}}$ & $\text{AssA}^{\texttt{R}}$ & \multirow{-3}{*}{\parbox[t]{0mm}{\tiny \rotatebox[origin=c]{90}{VTDA$_{\text{\textcolor{orange}{ass}}}$}}} & \multirow{-3}{*}{VTDA} \\ \midrule
ST Baselines$\dagger$ &  & 82.7 & 78.6 & 81.3 & 63.2 & 84.6 & \textbf{27.6} & 57.3 & 34.4 & 23.7 & 42.5 & 34.9 & 30.7 & 32.7 & 58.8 & 50.5 & 44.6 & 52.1 & 223.4 \\
\rowcolor{LightCyan} \textbf{\ourmodel} & \multirow{-2}{*}{ConvNeXt-T} & \textbf{83.2} & \textbf{80.0} & \textbf{82.1} & \textbf{64.8} & \textbf{86.3} & 26.7 & \textbf{57.9} & \textbf{36.0} & \textbf{28.4} & \textbf{42.8} & \textbf{36.2} & \textbf{33.4} & \textbf{34.8} & \textbf{60.4} & \textbf{52.1} & \textbf{45.3} & \textbf{53.5} & \textbf{228.3 (\textcolor{ao(english)}{+4.9})} \\ \midrule
ST Baselines$\dagger$ &  & 83.0 & 78.8 & 81.6 & 64.9 & 85.8 & \textbf{28.1} & \textbf{58.3} & 35.2 & 24.7 & \textbf{46.2} & 35.0 & 31.4 & 33.6 & 59.2 & 50.8 & 46.1 & 52.8 & 226.3 \\
\rowcolor{LightCyan} \textbf{\ourmodel} & \multirow{-2}{*}{ConvNeXt-B} & \textbf{83.3} & \textbf{80.0} & \textbf{82.2} & \textbf{65.9} & \textbf{86.0} & 27.1 & 58.1 & \textbf{36.3} & \textbf{29.4} & 45.5 & \textbf{37.6} & \textbf{34.7} & \textbf{36.0} & \textbf{60.8} & \textbf{51.9} & \textbf{48.3} & \textbf{54.3} & \textbf{230.7 (\textcolor{ao(english)}{+4.4})} \\ \bottomrule
\end{tabular}
\vspace{-0.1in}
\end{table*}
\setlength{\tabcolsep}{6pt}

\definecolor{LightGray}{rgb}{0.88,0.88,0.88}
\setlength{\tabcolsep}{3.8pt}
\begin{table*}[]
\scriptsize
\centering
\caption{
Comparison of \ourmodel to single-task models.
Our single-task baselines are highlighted in gray, which use the same task decoders as \ourmodelbase.
$\dagger$~indicates results from the official BDD100K model zoo and $\ddagger$~indicates results from prior published works.
}
\label{tab:sup-singletask}
\vspace{-0.14in}
\begin{tabular}{l|c|ccccccccccccccc}
\toprule
\multirow{2}{*}{Method} & \multirow{2}{*}{\begin{tabular}{@{}c@{}}Base\\Network\end{tabular}} & \multicolumn{2}{c}{Tagging} & Sem. & Driv. & Lane & Det. & Ins. & Pose & Flow & \multicolumn{3}{c}{MOT} & \multicolumn{3}{c}{MOTS} \\
 & & $\text{Acc}^{\texttt{Gw}}$ & $\text{Acc}^{\texttt{Gs}}$ & $\text{IoU}^{\texttt{S}}$ & $\text{IoU}^{\texttt{A}}$ & $\text{IoU}^{\texttt{L}}$ & $\text{AP}^{\texttt{D}}$ & $\text{AP}^{\texttt{I}}$ & $\text{AP}^{\texttt{P}}$ & $\text{IoU}^{\texttt{F}}$ & mMOTA & mIDF1 & mAP & mMOTSA & mIDF1 & mAP \\ \midrule
\textit{ResNet-50} & \\
\rowcolor{LightGray} ResNet~\cite{he2016deep} &  & 81.9 &	77.9 & \textcolor{gray}{–} & \textcolor{gray}{–} & \textcolor{gray}{–} & \textcolor{gray}{–} & \textcolor{gray}{–} & \textcolor{gray}{–} & \textcolor{gray}{–} & \textcolor{gray}{–} & \textcolor{gray}{–} & \textcolor{gray}{–} & \textcolor{gray}{–} & \textcolor{gray}{–} & \textcolor{gray}{–} \\
\rowcolor{LightGray} Semantic FPN~\cite{kirillov2019panoptic} &  & \textcolor{gray}{–} & \textcolor{gray}{–} & 59.7 & \textcolor{gray}{–} & \textcolor{gray}{–} & \textcolor{gray}{–} & \textcolor{gray}{–} & \textcolor{gray}{–} & \textcolor{gray}{–} & \textcolor{gray}{–} & \textcolor{gray}{–} & \textcolor{gray}{–} & \textcolor{gray}{–} & \textcolor{gray}{–} & \textcolor{gray}{–} \\
DNLNet~\cite{yin2020disentangled}$\dagger$ &  & \textcolor{gray}{–} & \textcolor{gray}{–} & 62.6 & \textcolor{gray}{–} & \textcolor{gray}{–} & \textcolor{gray}{–} & \textcolor{gray}{–} & \textcolor{gray}{–} & \textcolor{gray}{–} & \textcolor{gray}{–} & \textcolor{gray}{–} & \textcolor{gray}{–} & \textcolor{gray}{–} & \textcolor{gray}{–} & \textcolor{gray}{–} \\
DeepLabv3+~\cite{chen2018encoder}$\dagger$ &  & \textcolor{gray}{–} & \textcolor{gray}{–} & 64.0 & \textcolor{gray}{–} & \textcolor{gray}{–} & \textcolor{gray}{–} & \textcolor{gray}{–} & \textcolor{gray}{–} & \textcolor{gray}{–} & \textcolor{gray}{–} & \textcolor{gray}{–} & \textcolor{gray}{–} & \textcolor{gray}{–} & \textcolor{gray}{–} & \textcolor{gray}{–} \\
\rowcolor{LightGray} Semantic FPN~\cite{kirillov2019panoptic} &  & \textcolor{gray}{–} & \textcolor{gray}{–} & \textcolor{gray}{–} & 83.9 & \textcolor{gray}{–} & \textcolor{gray}{–} & \textcolor{gray}{–} & \textcolor{gray}{–} & \textcolor{gray}{–} & \textcolor{gray}{–} & \textcolor{gray}{–} & \textcolor{gray}{–} & \textcolor{gray}{–} & \textcolor{gray}{–} & \textcolor{gray}{–} \\
DeepLabv3+~\cite{chen2018encoder}$\dagger$ &  & \textcolor{gray}{–} & \textcolor{gray}{–} & \textcolor{gray}{–} & 84.4 & \textcolor{gray}{–} & \textcolor{gray}{–} & \textcolor{gray}{–} & \textcolor{gray}{–} & \textcolor{gray}{–} & \textcolor{gray}{–} & \textcolor{gray}{–} & \textcolor{gray}{–} & \textcolor{gray}{–} & \textcolor{gray}{–} & \textcolor{gray}{–} \\
DNLNet~\cite{yin2020disentangled}$\dagger$ &  & \textcolor{gray}{–} & \textcolor{gray}{–} & \textcolor{gray}{–} & 84.8 & \textcolor{gray}{–} & \textcolor{gray}{–} & \textcolor{gray}{–} & \textcolor{gray}{–} & \textcolor{gray}{–} & \textcolor{gray}{–} & \textcolor{gray}{–} & \textcolor{gray}{–} & \textcolor{gray}{–} & \textcolor{gray}{–} & \textcolor{gray}{–} \\
\rowcolor{LightGray} Semantic FPN~\cite{kirillov2019panoptic} &  & \textcolor{gray}{–} & \textcolor{gray}{–} & \textcolor{gray}{–} & \textcolor{gray}{–} & 28.4 & \textcolor{gray}{–} & \textcolor{gray}{–} & \textcolor{gray}{–} & \textcolor{gray}{–} & \textcolor{gray}{–} & \textcolor{gray}{–} & \textcolor{gray}{–} & \textcolor{gray}{–} & \textcolor{gray}{–} & \textcolor{gray}{–} \\
\rowcolor{LightGray} Faster R-CNN~\cite{ren2015faster} &  & \textcolor{gray}{–} & \textcolor{gray}{–} & \textcolor{gray}{–} & \textcolor{gray}{–} & \textcolor{gray}{–} & 32.3 & \textcolor{gray}{–} & \textcolor{gray}{–} & \textcolor{gray}{–} & \textcolor{gray}{–} & \textcolor{gray}{–} & \textcolor{gray}{–} & \textcolor{gray}{–} & \textcolor{gray}{–} & \textcolor{gray}{–} \\
Deform. DETR~\cite{zhu2020deformable}$\dagger$ &  & \textcolor{gray}{–} & \textcolor{gray}{–} & \textcolor{gray}{–} & \textcolor{gray}{–} & \textcolor{gray}{–} & 32.1 & \textcolor{gray}{–} & \textcolor{gray}{–} & \textcolor{gray}{–} & \textcolor{gray}{–} & \textcolor{gray}{–} & \textcolor{gray}{–} & \textcolor{gray}{–} & \textcolor{gray}{–} & \textcolor{gray}{–} \\
Cascade R-CNN~\cite{cai2018cascade}$\dagger$ &  & \textcolor{gray}{–} & \textcolor{gray}{–} & \textcolor{gray}{–} & \textcolor{gray}{–} & \textcolor{gray}{–} & 33.8 & \textcolor{gray}{–} & \textcolor{gray}{–} & \textcolor{gray}{–} & \textcolor{gray}{–} & \textcolor{gray}{–} & \textcolor{gray}{–} & \textcolor{gray}{–} & \textcolor{gray}{–} & \textcolor{gray}{–} \\
\rowcolor{LightGray} Mask R-CNN~\cite{ren2015faster} &  & \textcolor{gray}{–} & \textcolor{gray}{–} & \textcolor{gray}{–} & \textcolor{gray}{–} & \textcolor{gray}{–} & \textcolor{gray}{–} & 20.2 & \textcolor{gray}{–} & \textcolor{gray}{–} & \textcolor{gray}{–} & \textcolor{gray}{–} & \textcolor{gray}{–} & \textcolor{gray}{–} & \textcolor{gray}{–} & \textcolor{gray}{–} \\
Cascade R-CNN~\cite{ren2015faster}$\dagger$ &  & \textcolor{gray}{–} & \textcolor{gray}{–} & \textcolor{gray}{–} & \textcolor{gray}{–} & \textcolor{gray}{–} & \textcolor{gray}{–} & 21.4 & \textcolor{gray}{–} & \textcolor{gray}{–} & \textcolor{gray}{–} & \textcolor{gray}{–} & \textcolor{gray}{–} & \textcolor{gray}{–} & \textcolor{gray}{–} & \textcolor{gray}{–} \\
HTC~\cite{chen2019hybrid}$\dagger$ &  & \textcolor{gray}{–} & \textcolor{gray}{–} & \textcolor{gray}{–} & \textcolor{gray}{–} & \textcolor{gray}{–} & \textcolor{gray}{–} & 21.7 & \textcolor{gray}{–} & \textcolor{gray}{–} & \textcolor{gray}{–} & \textcolor{gray}{–} & \textcolor{gray}{–} & \textcolor{gray}{–} & \textcolor{gray}{–} & \textcolor{gray}{–} \\
\rowcolor{LightGray} Simple Baseline~\cite{Xiao_2018_ECCV} &  & \textcolor{gray}{–} & \textcolor{gray}{–} & \textcolor{gray}{–} & \textcolor{gray}{–} & \textcolor{gray}{–} & \textcolor{gray}{–} & \textcolor{gray}{–} & 37.0 & \textcolor{gray}{–} & \textcolor{gray}{–} & \textcolor{gray}{–} & \textcolor{gray}{–} & \textcolor{gray}{–} & \textcolor{gray}{–} & \textcolor{gray}{–} \\
\rowcolor{LightGray} PWC-Net~\cite{sun2018pwc} &  & \textcolor{gray}{–} & \textcolor{gray}{–} & \textcolor{gray}{–} & \textcolor{gray}{–} & \textcolor{gray}{–} & \textcolor{gray}{–} & \textcolor{gray}{–} & \textcolor{gray}{–} & 59.6 & \textcolor{gray}{–} & \textcolor{gray}{–} & \textcolor{gray}{–} & \textcolor{gray}{–} & \textcolor{gray}{–} & \textcolor{gray}{–} \\
\rowcolor{LightGray} QDTrack~\cite{pang2021quasi} &  & \textcolor{gray}{–} & \textcolor{gray}{–} & \textcolor{gray}{–} & \textcolor{gray}{–} & \textcolor{gray}{–} & \textcolor{gray}{–} & \textcolor{gray}{–} & \textcolor{gray}{–} & \textcolor{gray}{–} & 36.6 & 50.8 & 32.6 & \textcolor{gray}{–} & \textcolor{gray}{–} & \textcolor{gray}{–} \\
Unicorn~\cite{unicorn}$\ddagger$ &  & \textcolor{gray}{–} & \textcolor{gray}{–} & \textcolor{gray}{–} & \textcolor{gray}{–} & \textcolor{gray}{–} & \textcolor{gray}{–} & \textcolor{gray}{–} & \textcolor{gray}{–} & \textcolor{gray}{–} & 35.1 & \textcolor{gray}{–} & \textcolor{gray}{–} & \textcolor{gray}{–} & \textcolor{gray}{–} & \textcolor{gray}{–} \\
TETer~\cite{teter}$\ddagger$ &  & \textcolor{gray}{–} & \textcolor{gray}{–} & \textcolor{gray}{–} & \textcolor{gray}{–} & \textcolor{gray}{–} & \textcolor{gray}{–} & \textcolor{gray}{–} & \textcolor{gray}{–} & \textcolor{gray}{–} & 39.0 & 53.6 & \textcolor{gray}{–} & \textcolor{gray}{–} & \textcolor{gray}{–} & \textcolor{gray}{–} \\
\rowcolor{LightGray} QDTrack-MOTS~\cite{pang2021quasi} &  & \textcolor{gray}{–} & \textcolor{gray}{–} & \textcolor{gray}{–} & \textcolor{gray}{–} & \textcolor{gray}{–} & \textcolor{gray}{–} & \textcolor{gray}{–} & \textcolor{gray}{–} & \textcolor{gray}{–} & \textcolor{gray}{–} & \textcolor{gray}{–} & \textcolor{gray}{–} & 23.5 & 44.5 & 25.5 \\
PCAN~\cite{ke2021prototypical}$\ddagger$ &  & \textcolor{gray}{–} & \textcolor{gray}{–} & \textcolor{gray}{–} & \textcolor{gray}{–} & \textcolor{gray}{–} & \textcolor{gray}{–} & \textcolor{gray}{–} & \textcolor{gray}{–} & \textcolor{gray}{–} & \textcolor{gray}{–} & \textcolor{gray}{–} & \textcolor{gray}{–} & 27.4 & 45.1 & 26.6 \\
VMT~\cite{ke2022video}$\ddagger$ &  & \textcolor{gray}{–} & \textcolor{gray}{–} & \textcolor{gray}{–} & \textcolor{gray}{–} & \textcolor{gray}{–} & \textcolor{gray}{–} & \textcolor{gray}{–} & \textcolor{gray}{–} & \textcolor{gray}{–} & \textcolor{gray}{–} & \textcolor{gray}{–} & \textcolor{gray}{–} & 28.7 & 45.7 & 28.3 \\
Unicorn~\cite{unicorn}$\ddagger$ & \multirow{-24}{*}{ResNet-50} & \textcolor{gray}{–} & \textcolor{gray}{–} & \textcolor{gray}{–} & \textcolor{gray}{–} & \textcolor{gray}{–} & \textcolor{gray}{–} & \textcolor{gray}{–} & \textcolor{gray}{–} & \textcolor{gray}{–} & \textcolor{gray}{–} & \textcolor{gray}{–} & \textcolor{gray}{–} & 30.8 & \textcolor{gray}{–} & \textcolor{gray}{–} \\
\rowcolor{LightCyan} \textbf{\ourmodel} &  & 83.2 & 79.7 & 63.8 & 85.4 & 27.8 & 33.4 & 27.1 & 39.7 & 60.3 & 36.4 & 51.5 & 34.7 & 33.7 & 49.4 & 31.6 \\
\midrule
\textit{System-level Comparison} & \\
UPerNet~\cite{xiao2018unified}$\dagger$ & ConvNeXt-B & \textcolor{gray}{–} & \textcolor{gray}{–} & 67.3 & \textcolor{gray}{–} & \textcolor{gray}{–} & \textcolor{gray}{–} & \textcolor{gray}{–} & \textcolor{gray}{–} & \textcolor{gray}{–} & \textcolor{gray}{–} & \textcolor{gray}{–} & \textcolor{gray}{–} & \textcolor{gray}{–} & \textcolor{gray}{–} & \textcolor{gray}{–} \\
AFA-DLA~\cite{yang2023dense}$\ddagger$ & DLA-169 & \textcolor{gray}{–} & \textcolor{gray}{–} & 67.5 & \textcolor{gray}{–} & \textcolor{gray}{–} & \textcolor{gray}{–} & \textcolor{gray}{–} & \textcolor{gray}{–} & \textcolor{gray}{–} & \textcolor{gray}{–} & \textcolor{gray}{–} & \textcolor{gray}{–} & \textcolor{gray}{–} & \textcolor{gray}{–} & \textcolor{gray}{–} \\
Cascade R-CNN~\cite{cai2018cascade}$\dagger$ & ConvNeXt-B & \textcolor{gray}{–} & \textcolor{gray}{–} & \textcolor{gray}{–} & \textcolor{gray}{–} & \textcolor{gray}{–} & 36.2 & \textcolor{gray}{–} & \textcolor{gray}{–} & \textcolor{gray}{–} & \textcolor{gray}{–} & \textcolor{gray}{–} & \textcolor{gray}{–} & \textcolor{gray}{–} & \textcolor{gray}{–} & \textcolor{gray}{–} \\
Mask Transfiner~\cite{transfiner}$\ddagger$ & ResNet-101 & \textcolor{gray}{–} & \textcolor{gray}{–} & \textcolor{gray}{–} & \textcolor{gray}{–} & \textcolor{gray}{–} & \textcolor{gray}{–} & 23.6 & \textcolor{gray}{–} & \textcolor{gray}{–} & \textcolor{gray}{–} & \textcolor{gray}{–} & \textcolor{gray}{–} & \textcolor{gray}{–} & \textcolor{gray}{–} & \textcolor{gray}{–} \\
Cascade R-CNN~\cite{cai2018cascade}$\dagger$ & ConvNeXt-B & \textcolor{gray}{–} & \textcolor{gray}{–} & \textcolor{gray}{–} & \textcolor{gray}{–} & \textcolor{gray}{–} & \textcolor{gray}{–} & 27.5 & \textcolor{gray}{–} & \textcolor{gray}{–} & \textcolor{gray}{–} & \textcolor{gray}{–} & \textcolor{gray}{–} & \textcolor{gray}{–} & \textcolor{gray}{–} & \textcolor{gray}{–} \\
Unicorn~\cite{unicorn}$\ddagger$ & ConvNeXt-L & \textcolor{gray}{–} & \textcolor{gray}{–} & \textcolor{gray}{–} & \textcolor{gray}{–} & \textcolor{gray}{–} & \textcolor{gray}{–} & \textcolor{gray}{–} & \textcolor{gray}{–} & \textcolor{gray}{–} & 41.2 & 54.0 & – & \textcolor{gray}{–} & \textcolor{gray}{–} & \textcolor{gray}{–} \\
Unicorn~\cite{unicorn}$\ddagger$ & ConvNeXt-L & \textcolor{gray}{–} & \textcolor{gray}{–} & \textcolor{gray}{–} & \textcolor{gray}{–} & \textcolor{gray}{–} & \textcolor{gray}{–} & \textcolor{gray}{–} & \textcolor{gray}{–} & \textcolor{gray}{–} & \textcolor{gray}{–} & \textcolor{gray}{–} & \textcolor{gray}{–} & 29.6 & 44.2 & – \\
\rowcolor{LightCyan} \textbf{\ourmodel} & ConvNeXt-B & 83.3 & 80.0 & 65.9 & 86.0 & 27.1 & 36.3 & 29.4 & 45.5 & 60.8 & 38.6 & 55.2 & 37.6 & 35.3 & 52.6 & 34.7 \\
\bottomrule
\end{tabular}
\vspace{-0.12in}
\end{table*}
\setlength{\tabcolsep}{6pt}

\section{Additional Base Networks} \label{sec:sup-basenetwork}
We provide comparisons of \ourmodel against single-task baselines across additional base networks in \tableautorefname~\ref{tab:sup-backbones}, including ConvNeXt-T and ConvNeXt-B~\cite{liu2022convnet}.
\ourmodelbase maintains its advantage in overall performance across all base networks and achieves performance gains across most tasks.
In particular, localization performance improves significantly as model capacity increases, reaching a high score of 36.3 detection AP and 29.4 instance segmentation AP (ConvNeXt-B).
This also translates to 37.6 MOT AP and 34.7 MOTS AP.
Furthermore, semantic segmentation also observes consistent improvements up to 65.9 mIoU.
However, we observe that drivable area and lane detection do not noticeably benefit from the increased capacity, maintaining a similar performance across all base networks.

\section{Full Single-Task Comparison} \label{sec:sup-singletask}
We compare \ourmodelbase against single-task baselines, models from the official BDD100K model zoo\protect\footnotemark, and state-of-the-art methods across all tasks.
For MOT and MOTS, we use the metrics used by the official benchmarks, mMOTA~\cite{stief2007clear} and mIDF1~\cite{ristani2016performance}.
All other tasks use the same metrics as in \ourbenchmark.
The results are shown in \tableautorefname~\ref{tab:sup-singletask}.
\footnotetext{\url{https://github.com/SysCV/bdd100k-models}}

\ourmodel can achieve SOTA performance on several benchmarks and competitive performance on the rest, despite using only a single model for all tasks with much simpler task-specific heads.
For the ResNet-50 comparison,
\ourmodel achieves significantly better performance on MOTS compared to VMT~\cite{ke2022video}, obtaining an improvement of 5 points in mMOTSA, 3.7 points in mIDF1, and 2.7 points in mAP without using extra tracking annotations or specialized modules.
On instance segmentation, \ourmodel obtains an improvement of 5.4 points over HTC~\cite{chen2019hybrid}, which utilizes a complex cascade structure.
On semantic segmentation and drivable area segmentation, \ourmodel achieves competitive performance with DeepLabv3+~\cite{chen2018encoder}, despite only using a simple FPN structure.

We also provide system-level comparisons with SOTA methods on established BDD100K benchmarks.
By simply scaling up the capacity of the base network, \ourmodel achieves performance gains across the board, outperforming other methods that utilize hand-designed task-specific modules with a simple unified structure.
On semantic segmentation and object detection, \ourmodel again achieves competitive performance with specialized models.
On instance segmentation, \ourmodel achieves an improvement of 1.9 points in AP over Cascade R-CNN~\cite{cai2018cascade} that uses a cascade structure for mask refinement.
On MOT and MOTS, \ourmodelbase obtains significantly higher mIDF1 of 1.2 points for MOT and 8.4 points for MOTS over Unicorn~\cite{unicorn}, demonstrating that it is much better at object association.
However, since Unicorn uses a stronger detector, larger base network, and additional tracking training data, it achieves better mMOTA in MOT.
Nevertheless, \ourmodel still obtains an improvement of 5.7 points in mMOTSA in MOTS.

\setlength{\tabcolsep}{8pt}
\begin{table*}[t]
\footnotesize
\centering
\caption{
Statistics of tasks and available annotations in KITTI~\cite{Geiger2012CVPR}.
}
\label{tab:sup-kitti-splits}
\vspace{-0.14in}
\begin{tabular}{l|cc|c}
\toprule
Set & Images (Train / Val) & \% Total Images & Tasks with Annotations \\ \midrule
Detection & 6.2K / 1.3K & 54\% & Image tagging, object detection \\
Segmentation & 160 / 40 & 1\% & Instance segmentation, semantic segmentation \\
Drivable & 236 / 53 & 2\% & Drivable area segmentation \\
Tracking & 5K / 3K (=12 / 9 videos) & 43\% & MOT, MOTS, semantic segmentation \\
\bottomrule
\end{tabular}
\vspace{-0.12in}
\end{table*}

\setlength{\tabcolsep}{6.7pt}
\begin{table*}[]
\centering
\scriptsize
\caption{
Comparison of \ourmodelbase against single-task (ST) and multi-task (MT) baselines on KITTI validation set.
CPF denotes our training protocol, and $\dagger$~denotes a separate model is trained for each task.
\ourmodelbase achieves significantly better \ourmetric than both ST and MT baselines.
\textbf{Black} / \textit{\textcolor{blue}{blue}} indicate best / second best.
}
\label{tab:sup-kitti}
\vspace{-0.14in}
\begin{tabular}{lc|cc|ccc|ccccc|ccc|c}
\toprule
\multirow{3}{*}{Method} & \multirow{3}{*}{CPF} & Class. &  & \multicolumn{2}{c}{Segmentation} &  & \multicolumn{4}{c}{Localization} &  & \multicolumn{2}{c}{Association} &  &  \\
 &  & Tag &  & Sem. & Driv. &  & Det. & Ins. & MOT & MOTS &  & MOT & MOTS &  &  \\
 & & $\text{Acc}^{\texttt{G}}$ & \multirow{-3}{*}{\parbox[t]{1.5mm}{\tiny \rotatebox[origin=c]{90}{VTDA$_{\text{\textcolor{ao(english)}{cls}}}$}}} & $\text{IoU}^{\texttt{S}}$ & $\text{IoU}^{\texttt{A}}$ & \multirow{-3}{*}{\parbox[t]{0mm}{\tiny \rotatebox[origin=c]{90}{VTDA$_{\text{\textcolor{amber}{seg}}}$}}} & $\text{AP}^{\texttt{D}}$ & $\text{AP}^{\texttt{I}}$ & $\text{AP}^{\texttt{T}}$ & $\text{AP}^{\texttt{R}}$ & \multirow{-3}{*}{\parbox[t]{0mm}{\tiny \rotatebox[origin=c]{90}{VTDA$_{\text{\textcolor{ceruleanblue}{loc}}}$}}} & $\text{AssA}^{\texttt{T}}$ & $\text{AssA}^{\texttt{R}}$ & \multirow{-3}{*}{\parbox[t]{0mm}{\tiny \rotatebox[origin=c]{90}{VTDA$_{\text{\textcolor{orange}{ass}}}$}}} & \multirow{-3}{*}{VTDA} \\ \midrule
ST Baselines$\dagger$ & & \multicolumn{2}{c|}{49.7} & 48.0 & 96.8 & 72.4 & \textbf{60.5} & 24.9 & 48.4 & 47.1 & 45.2 & {\color[HTML]{0000FF} \textit{61.8}} & 61.4 & 61.6 & 228.9 \\
MT Baseline & & \multicolumn{2}{c|}{{\color[HTML]{0000FF} \textit{50.1}}} & 42.9 & 94.9 & 68.9 & 53.4 & {\color[HTML]{0000FF} \textit{31.3}} & {\color[HTML]{0000FF} \textit{51.4}} & 52.3 & 47.1 & 60.9 & 61.6 & 61.3 & 227.4 (\textcolor{purple}{-1.5}) \\
\rowcolor{LightCyan}  & \textcolor{lightgray}{\xmark} & \multicolumn{2}{c|}{\textbf{50.2}} & \textbf{51.3} & \textbf{97.5} & \textbf{74.4} & {\color[HTML]{0000FF} \textit{56.2}} & 31.2 & 49.2 & {\color[HTML]{0000FF} \textit{53.3}} & {\color[HTML]{0000FF} \textit{47.5}} & 61.1 & {\color[HTML]{0000FF} \textit{62.8}} & {\color[HTML]{0000FF} \textit{62.0}} & {\color[HTML]{0000FF} \textit{234.0}} (\textcolor{ao(english)}{+5.1}) \\
\rowcolor{LightCyan} \multirow{-2}{*}{\textbf{\ourmodel}} & \textcolor{ao(english)}{\cmark} & \multicolumn{2}{c|}{{\color[HTML]{0000FF} \textit{50.1}}} & {\color[HTML]{0000FF} \textit{50.2}} & {\color[HTML]{0000FF} \textit{97.4}} & {\color[HTML]{0000FF} \textit{73.8}} & 52.1 & \textbf{33.5} & \textbf{54.8} & \textbf{55.0} & \textbf{48.9} & \textbf{65.3} & \textbf{66.0} & \textbf{65.7} & \textbf{238.4 (\textcolor{ao(english)}{+9.5})} \\
\bottomrule
\end{tabular}
\vspace{-0.12in}
\end{table*}
\setlength{\tabcolsep}{6pt}

\section{Experiments on KITTI Dataset} \label{sec:sup-kitti}
We conduct additional experiments on the KITTI dataset~\cite{Geiger2012CVPR} to demonstrate the versatility of our proposed network and training protocol.

\parsection{Dataset.}
KITTI is a real-world, autonomous driving benchmark suite that contains various vision tasks, including object detection, tracking, and segmentation.
Compared to BDD100K~\cite{yu2020bdd100k}, KITTI is much smaller in scale ($\sim$10\% of data) and uses fewer object categories (mainly pedestrians and cars).
As only the training set is publicly available, we split the training set into train and validation sets for our experiments.
Similar to BDD100K, the annotation density of each image set varies widely,
and the statistics are shown in \tableautorefname~\ref{tab:sup-kitti-splits}.

\parsection{Evaluation Setting.}
We construct a similar heterogeneous multi-task setup on KITTI for training and evaluation.
We use seven tasks that are consistent with \ourbenchmark: image tagging, drivable area, semantic/instance segmentation, object detection, and MOT/MOTS.
To compute \ourmetric, we simply drop the missing tasks from the averages.
As we do not have estimates of task sensitivities across different base networks, we do not scale each task score and opt for a simple average.

\parsection{Comparison to Baselines.}
We perform the same comparison to single-task (ST) and multi-task (MT) baselines, and we use the same architecture as with \ourbenchmark but removing the missing tasks' decoders.
As KITTI exhibits the same data imbalance issue, we use the same \ourlearn training protocol as on \ourbenchmark, but we do not find pseudo-labels to be necessary.
The results are shown in \tableautorefname~\ref{tab:sup-kitti}.

Compared to the single-task baselines, the multi-task baseline that jointly optimizes all tasks achieves worse performance on most tasks, demonstrating that a naive architecture and training protocol are not sufficient.
\ourmodel improves performance of most tasks and leads to much better segmentation and localization scores.
With a better training protocol, \ourlearn enables \ourmodel to achieve significantly better performance on most tasks and an improvement of 9.5 points in \ourmetric.
However, we note that there exists negative transfer between object detection, instance segmentation, and MOT localization on KITTI, \ie, improvement in one score leads to a drop in the others.
We attribute this to differences in annotation between each image set, as KITTI is not designed for multi-task learning.
Nevertheless, these results demonstrate the generalizability and effectiveness of our proposed network and training protocol.

\section{Additional Ablation Studies} \label{sec:sup-abs}
We provide additional ablation studies on components of our optimization strategy.
In these experiments, we use~\ourmodelbase with ResNet-50~\cite{he2016deep} and use the default training parameters with~\ourlearn, unless otherwise stated.

\parsection{Loss Weights.}
We provide full results of \ourmodelbase with different loss weights configurations in
\tableautorefname~\ref{tab:sup-lossw}
to complement \figureautorefname~6 in the main paper.
For each configuration, we show the loss weights of each task in the first row and the task scores in the second row,
and we increase the loss weights of a particular group of tasks to boost the performance of the network in that aspect.
Doing so consistently improves the performance in each aspect at the cost of a drop in performance in other aspects.
This enables prioritization of certain tasks over others through the choice of loss weights.
The overall performance of \ourmodelbase remains stable across all configurations.

\setlength{\tabcolsep}{4.4pt}
\begin{table*}[]
\centering
\scriptsize
\caption{
Ablation study on loss weights with~\ourmodelbase on \ourbenchmark validation set.
We show loss weights for each task (first row) and task-specific performance along with \ourmetric (second row).
For loss weights, differences between settings are \underline{underlined}.
Increasing loss weights on a group of tasks boosts the performance of \ourmodelbase in that aspect, enabling prioritization of task performance depending on application.
}
\label{tab:sup-lossw}
\vspace{-0.14in}
\begin{tabular}{c|ccc|cccc|cccccc|cccc|c}
\toprule
\multirow{3}{*}{Loss weights} & \multicolumn{2}{c}{Classification} &  & \multicolumn{3}{c}{Segmentation} &  & \multicolumn{5}{c}{Localization} &  & \multicolumn{3}{c}{Association} &  &  \\
 & \multicolumn{2}{c}{Tagging} &  & Sem. & Driv. & Lane &  & Det. & Ins. & Pose & MOT & MOTS &  & Flow & MOT & MOTS &  &  \\
 & $\text{Acc}^{\texttt{Gw}}$ & $\text{Acc}^{\texttt{Gs}}$ & \multirow{-3}{*}{\parbox[t]{1.5mm}{\tiny \rotatebox[origin=c]{90}{VTDA$_{\text{\textcolor{ao(english)}{cls}}}$}}} & $\text{IoU}^{\texttt{S}}$ & $\text{IoU}^{\texttt{A}}$ & $\text{IoU}^{\texttt{L}}$ & \multirow{-3}{*}{\parbox[t]{0mm}{\tiny \rotatebox[origin=c]{90}{VTDA$_{\text{\textcolor{amber}{seg}}}$}}} & $\text{AP}^{\texttt{D}}$ & $\text{AP}^{\texttt{I}}$ & $\text{AP}^{\texttt{P}}$ & $\text{AP}^{\texttt{T}}$ & $\text{AP}^{\texttt{R}}$ & \multirow{-3}{*}{\parbox[t]{0mm}{\tiny \rotatebox[origin=c]{90}{VTDA$_{\text{\textcolor{ceruleanblue}{loc}}}$}}} & $\text{IoU}^{\texttt{F}}$ & $\text{AssA}^{\texttt{T}}$ & $\text{AssA}^{\texttt{R}}$ & \multirow{-3}{*}{\parbox[t]{0mm}{\tiny \rotatebox[origin=c]{90}{VTDA$_{\text{\textcolor{orange}{ass}}}$}}} & \multirow{-3}{*}{VTDA} \\ \midrule
Default & \textit{0.05} & \textit{0.05} &  & \textit{1.0} & \textit{1.0} & \textit{2.0} &  & \textit{2.0} & \textit{2.0} & \textit{800.0} & \textit{1.0} &  &  & \textit{1.0} & \textit{1.0} &  &  &  \\
\rowcolor{LightCyan}  & 83.2 & 79.7 & 82.0 & 63.8 & 85.4 & 27.8 & 57.8 & 33.4 & 27.1 & 39.7 & 34.7 & 31.6 & 32.9 & 60.3 & 50.1 & 45.1 & 52.7 & \textbf{225.3} \\ \midrule
$2\lambda_G$ & \underline{\textit{0.1}} & \underline{\textit{0.1}} &  & \textit{1.0} & \textit{1.0} & \textit{2.0} &  & \textit{2.0} & \textit{2.0} & \textit{800.0} & \textit{1.0} &  &  & \textit{1.0} & \textit{1.0} &  &  &  \\
\rowcolor{LightCyan} & \textbf{83.4} & \textbf{79.9} & \textbf{82.2} & 63.7 & 85.4 & 27.4 & 57.6 & 33.1 & 26.6 & 39.4 & 34.0 & 31.0 & 32.4 & 60.2 & 49.9 & 44.9 & 52.5 & 224.7 \\ \midrule
$2\lambda_S,2\lambda_A,2\lambda_L$ & \textit{0.05} & \textit{0.05} &  & \underline{\textit{2.0}} & \underline{\textit{2.0}} & \underline{\textit{4.0}} &  & \textit{2.0} & \textit{2.0} & \textit{800.0} & \textit{1.0} &  &  & \textit{1.0} & \textit{1.0} &  &  &  \\
\rowcolor{LightCyan} & 83.1 & 79.6 & 81.9 & \textbf{64.0} & \textbf{85.5} & \textbf{28.0} & \textbf{58.0} & 33.2 & 27.0 & 38.9 & 33.9 & 31.4 & 32.5 & 60.1 & 49.6 & 44.8 & 52.3 & 224.7 \\ \midrule
$2\lambda_D,2\lambda_I,2\lambda_P$ & \textit{0.05} & \textit{0.05} &  & \textit{1.0} & \textit{1.0} & \textit{2.0} &  & \underline{\textit{4.0}} & \underline{\textit{4.0}} & \underline{\textit{1600.0}} & \textit{1.0} &  &  & \textit{1.0} & \textit{1.0} &  &  &  \\
\rowcolor{LightCyan} & 83.3 & 79.7 & 82.1 & 63.4 & 85.1 & 27.0 & 57.3 & \textbf{33.6} & \textbf{27.4} & \textbf{40.4} & \textbf{35.0} & \textbf{31.8} & \textbf{33.2} & 60.0 & 49.9 & 45.0 & 52.5 & 225.0 \\ \midrule
$2\lambda_F,2\lambda_T$ & \textit{0.05} & \textit{0.05} &  & \textit{1.0} & \textit{1.0} & \textit{2.0} &  & \textit{2.0} & \textit{2.0} & \textit{800.0} & \underline{\textit{2.0}} &  &  & \underline{\textit{2.0}} & \underline{\textit{2.0}} &  &  &  \\
\rowcolor{LightCyan} & 83.3 & 79.8 & 82.1 & 63.6 & 85.1 & 27.2 & 57.4 & 33.3 & 27.1 & 39.6 & 34.6 & 31.6 & 32.8 & \textbf{60.6} & \textbf{50.4} & \textbf{45.4} & \textbf{53.0} & \textbf{225.3} \\
\bottomrule
\end{tabular}
\vspace{-0.12in}
\end{table*}
\setlength{\tabcolsep}{6pt}

\parsection{Data Sampling Strategies.}
We additionally investigate different data sampling strategies used during joint optimization on all \ourbenchmark tasks.
Our default strategy, set-level round-robin, samples a batch of training data from each image set in order (\sectionautorefname~\ref{sec:sup-curr}).
We also experiment with not using any sampling (None), uniform sampling (Uniform), and weighted sampling (Weighted), following~\cite{likhosherstov2021polyvit}.
Uniform and Weighted use a stochastic schedule that samples from a uniform and a weighted distribution (proportional to size of each image set).
The results are shown in \tableautorefname~\ref{tab:sup-strats}.
We find that not using any sampling strategy achieves decent performance already and using a stochastic sampler does not achieve further performance gains.
This is due to data-imbalance, and the aforementioned strategies are biased towards one image set over another.
On the contrary, round-robin sampling better balances the data and obtains the best performance overall.

\setlength{\tabcolsep}{4.83pt}
\begin{table*}[]
\centering
\scriptsize
\caption{
Ablation study on data sampling strategies during joint training with~\ourmodelbase on \ourbenchmark validation set.
}
\label{tab:sup-strats}
\vspace{-0.14in}
\begin{tabular}{c|ccc|cccc|cccccc|cccc|c}
\toprule
\multirow{3}{*}{Strategy} & \multicolumn{2}{c}{Classification} &  & \multicolumn{3}{c}{Segmentation} &  & \multicolumn{5}{c}{Localization} &  & \multicolumn{3}{c}{Association} &  &  \\
 & \multicolumn{2}{c}{Tagging} &  & Sem. & Driv. & Lane &  & Det. & Ins. & Pose & MOT & MOTS &  & Flow & MOT & MOTS &  &  \\
 & $\text{Acc}^{\texttt{Gw}}$ & $\text{Acc}^{\texttt{Gs}}$ & \multirow{-3}{*}{\parbox[t]{1.5mm}{\tiny \rotatebox[origin=c]{90}{VTDA$_{\text{\textcolor{ao(english)}{cls}}}$}}} & $\text{IoU}^{\texttt{S}}$ & $\text{IoU}^{\texttt{A}}$ & $\text{IoU}^{\texttt{L}}$ & \multirow{-3}{*}{\parbox[t]{0mm}{\tiny \rotatebox[origin=c]{90}{VTDA$_{\text{\textcolor{amber}{seg}}}$}}} & $\text{AP}^{\texttt{D}}$ & $\text{AP}^{\texttt{I}}$ & $\text{AP}^{\texttt{P}}$ & $\text{AP}^{\texttt{T}}$ & $\text{AP}^{\texttt{R}}$ & \multirow{-3}{*}{\parbox[t]{0mm}{\tiny \rotatebox[origin=c]{90}{VTDA$_{\text{\textcolor{ceruleanblue}{loc}}}$}}} & $\text{IoU}^{\texttt{F}}$ & $\text{AssA}^{\texttt{T}}$ & $\text{AssA}^{\texttt{R}}$ & \multirow{-3}{*}{\parbox[t]{0mm}{\tiny \rotatebox[origin=c]{90}{VTDA$_{\text{\textcolor{orange}{ass}}}$}}} & \multirow{-3}{*}{VTDA} \\ \midrule
None & \textbf{83.3} & \textbf{79.9} & \textbf{82.2} & \textbf{64.1} & 84.8 & 27.2 & 57.4 & 33.0 & 26.5 & 39.1 & 34.2 & 31.4 & 32.4 & \textbf{60.4} & 49.4 & 44.8 & 52.4 & 224.3 \\
Uniform & 83.1 & 79.6 & 81.9 & 62.6 & 85.1 & 27.4 & 57.3 & 33.3 & \textbf{27.5} & 39.3 & \textbf{34.8} & 31.0 & 32.8 & 60.2 & \textbf{50.5} & 43.8 & 52.5 & 224.5 \\
Weighted & 83.2 & 79.7 & 82.0 & 62.6 & 84.9 & 27.7 & 57.4 & \textbf{33.6} & 27.0 & 39.6 & 34.4 & 31.3 & 32.8 & 60.2 & 49.7 & 44.4 & 52.3 & 224.4 \\
\midrule
\rowcolor{LightCyan} Round-robin & 83.2 & 79.7 & 82.0 & 63.8 & \textbf{85.4} & \textbf{27.8} & \textbf{57.8} & 33.4 & 27.1 & \textbf{39.7} & 34.7 & \textbf{31.6} & \textbf{32.9} & 60.3 & 50.1 & \textbf{45.1} & \textbf{52.7} & \textbf{225.3} \\
\bottomrule
\end{tabular}
\vspace{-0.12in}
\end{table*}
\setlength{\tabcolsep}{6pt}

\section{Compute Resource Comparison} \label{sec:sup-res}
We provide the full compute resource usage during inference comparison of \ourmodelbase, single-task, and multi-task baselines with the ResNet-50 base network in \tableautorefname~\ref{tab:sup-params} to complement Figure 5 in the main paper.
We show the number of model parameters, number of multiply-accumulate operations (MACs), and number of floating-point operations (FLOPs).
These are measured during model inference on a single GeForce RTX 2080 Ti GPU.
The total resource utilization of \ourmodelbase is less than that of the semantic segmentation baseline plus the MOTS baseline, showing that a unified network can drastically save computation.
By sharing a majority of the network, \ourmodelbase achieves much better computational efficiency compared to single-task baselines by tackling all ten tasks with only a single set of weights.
Additionally, \ourmodelbase only uses marginally more computation compared to the multi-task baseline, while achieving much better performance.

\setlength{\tabcolsep}{8pt}
\begin{table}[]
\footnotesize
\centering
\caption{
Full resource usage comparison during inference between \ourmodelbase, single-task (ST), and multi-task (MT) baselines.
\ourmodelbase achieves much better computational efficiency compared to single-task baselines.
}
\label{tab:sup-params}
\vspace{-0.14in}
\begin{tabular}{l|ccc}
\toprule
Model & Params (M) & MACs (G) & FLOPs (G) \\ \midrule
Tagging & 23.5 & 33.4 & 67.0 \\
Detection & 41.2 & 190.6 & 381.9 \\
Instance Seg. & 43.8 & 192.5 & 385.8 \\
Pose Estimation & 44.3 & 192.5 & 385.7 \\
Semantic Seg. & 28.6 & 133.5 & 267.4 \\
Drivable Area & 28.6 & 133.4 & 273.1 \\
Lane Estimation & 28.6 & 133.5 & 273.1 \\
Optical Flow & 5.7 & 166.8 & 334.8 \\
MOT & 56.6 & 192.2 & 385.2 \\
MOTS & 59.3 & 216.7 & 434.2 \\ \midrule
ST Sum & 360.1 & 1585.1 & 3189.1 \\
MT Baseline & 73.4 & 292.1 & 586.5 \\
\midrule
\rowcolor{LightCyan} \textbf{\ourmodelbase} & 73.3 & 309.9 & 622.1 \\ \bottomrule
\end{tabular}
\vspace{-0.12in}
\end{table}

\setlength{\tabcolsep}{4.3pt}
\begin{table*}[]
\scriptsize
\centering
\caption{
Analysis of \ourmetric with \ourmodelbase using ResNet-50 base network on VTD validation set.
$\dagger$~denotes a separate model is trained for each task.
We also show absolute and scaled differences in task-specific performance.
\ourmetric better balances the contribution of each task score, leading to a more informative metric for our setting.
}
\label{tab:sup-metric}
\vspace{-0.14in}
\begin{tabular}{c|ccc|cccc|cccccc|cccc|c}
\toprule
\multirow{3}{*}{Method} & \multicolumn{2}{c}{Classification} &  & \multicolumn{3}{c}{Segmentation} &  & \multicolumn{5}{c}{Localization} &  & \multicolumn{3}{c}{Association} &  &  \\
 & \multicolumn{2}{c}{Tagging} &  & Sem. & Driv. & Lane &  & Det. & Ins. & Pose & MOT & MOTS &  & Flow & MOT & MOTS &  &  \\
 & $\text{Acc}^{\texttt{Gw}}$ & $\text{Acc}^{\texttt{Gs}}$ & \multirow{-3}{*}{\parbox[t]{1.5mm}{\tiny \rotatebox[origin=c]{90}{VTDA$_{\text{\textcolor{ao(english)}{cls}}}$}}} & $\text{IoU}^{\texttt{S}}$ & $\text{IoU}^{\texttt{A}}$ & $\text{IoU}^{\texttt{L}}$ & \multirow{-3}{*}{\parbox[t]{0mm}{\tiny \rotatebox[origin=c]{90}{VTDA$_{\text{\textcolor{amber}{seg}}}$}}} & $\text{AP}^{\texttt{D}}$ & $\text{AP}^{\texttt{I}}$ & $\text{AP}^{\texttt{P}}$ & $\text{AP}^{\texttt{T}}$ & $\text{AP}^{\texttt{R}}$ & \multirow{-3}{*}{\parbox[t]{0mm}{\tiny \rotatebox[origin=c]{90}{VTDA$_{\text{\textcolor{ceruleanblue}{loc}}}$}}} & $\text{IoU}^{\texttt{F}}$ & $\text{AssA}^{\texttt{T}}$ & $\text{AssA}^{\texttt{R}}$ & \multirow{-3}{*}{\parbox[t]{0mm}{\tiny \rotatebox[origin=c]{90}{VTDA$_{\text{\textcolor{orange}{ass}}}$}}} & \multirow{-3}{*}{VTDA} \\ \midrule
ST Baselines$\dagger$ & 81.9 & 77.9 & 80.6 & 59.7 & 83.9 & \textbf{28.4} & 56.7 & 32.3 & 20.2 & 37.0 & 32.9 & 27.2 & 29.7 & 59.6 & 48.8 & 42.4 & 51.3 & 218.2 \\
\rowcolor{LightCyan} \textbf{\ourmodelbase} & \textbf{83.2} & \textbf{79.7} & \textbf{82.0} & \textbf{63.8} & \textbf{85.4} & 27.8 & \textbf{57.8} & \textbf{33.4} & \textbf{27.1} & \textbf{39.7} & \textbf{34.7} & \textbf{31.6} & \textbf{32.9} & \textbf{60.3} & \textbf{50.1} & \textbf{45.1} & \textbf{52.7} & \textbf{225.3 (\textcolor{ao(english)}{+7.1})} \\
\midrule
Absolute $\Delta$ & 1.3 & 1.8 & 1.6 & 4.1 & 1.5 & -0.6 & 1.7 & 1.1 & 6.9 & 2.7 & 1.8 & 4.4 & 3.4 & 0.7 & 1.3 & 2.7 & 1.6 & – \\
$\sigma_t$ & 0.4 & 0.6 & – & 2.0 & 0.7 & 0.9 & – & 1.1 & 1.7 & 3.1 & 1.0 & 1.7 & – & 0.9 & 0.8 & 1.4 & – & – \\
$s_t$ & 1.00 & 0.50 & – & 0.20 & 0.50 & 0.50 & – & 0.33 & 0.25 & 0.14 & 0.33 & 0.25 & – & 0.50 & 0.50 & 0.33 & – & – \\
Scaled $\Delta$ & 1.3 & 0.9 & 1.4 & 0.8 & 0.8 & -0.3 & 1.1 & 0.4 & 1.7 & 0.4 & 0.6 & 1.1 & 3.2 & 0.3 & 0.7 & 0.9 & 1.4 & – \\
\bottomrule
\end{tabular}
\vspace{-0.12in}
\end{table*}
\setlength{\tabcolsep}{6pt}

\section{VTD Challenge Details} \label{sec:sup-benchmark}
We present further details regarding our proposed \ourbenchmark challenge, detailing each task and our \ourmetric metric.

\subsection{Tasks}
We first detail each task based on its definition in BDD100K~\cite{yu2020bdd100k} and modifications made to build our \ourbenchmark challenge.

\parsection{Image Tagging.}
There are two classification tasks, weather and scene classification.
The weather conditions are rainy, snowy, clear, overcast, partly cloudy, and foggy (plus undefined).
The scene types are tunnel, residential, parking lot, city street, gas stations, and highway (plus undefined).

\parsection{Object Detection.}
BDD100K provides ten categories for detection: pedestrian, rider, car, truck, bus, train, motorcycle, bicycle, traffic light, and traffic sign.
To be consistent with the segmentation and tracking sets, we only use the first eight categories for detection and treat the final two as stuff (background) categories.

\parsection{Pose Estimation.}
Pedestrians and riders in BDD100K are labeled with 18 joint keypoints throughout the body.

\parsection{Drivable Area Segmentation.}
For drivable area segmentation, the prediction of background is important to account for regions of the image that are not drivable.
Thus, the network needs to predict three classes.
Accuracy of the background prediction is not considered in the final score.

\parsection{Lane Detection.}
Lanes in BDD100K are labeled with eight categories
and two attributes,
direction and style.
Categories include road curb, crosswalk, double white, double yellow, double other color, single white, single yellow, and single other color.
Directions include parallel and vertical, and styles include solid and dashed.
Thus, each lane has three different labels.
We treat lane detection as a contour detection problem
for each of the three labels.
During evaluation, the performance is averaged over the three labels.
Similar to drivable area, background pixels are also required for prediction but not considered for evaluation.
Before computing mIoU, we dilate the ground truth by five pixels to account for ambiguity during annotation.
Due to the slow speed of computation, we use a subsample of 1K images for evaluation.

\parsection{Semantic / Instance Segmentation.}
BDD100K has 19 categories for semantic segmentation, split into 8 thing (foreground) and 11 stuff categories.
The 8 thing categories are consistent with the detection set.
The 11 stuff categories include road, sidewalk, building, wall, fence, pole, traffic light, traffic sign, vegetation, terrain, and sky.
Thing masks also include instance information used for instance segmentation.

\parsection{MOT / MOTS.}
The object tracking categories are pedestrian, rider, car, truck, bus, train, motorcycle, and bicycle.

\parsection{Optical Flow Estimation.}
We use a proxy evaluation method based on MOTS labels to evaluate optical flow estimation.
Given segmentation masks of two consecutive frames $M_t$, $M_{t-1} \in \mathbb{R}^{H \times W}$
and the predicted optical flow \mbox{$V \in \mathbb{R}^{H \times W \times 2}$}, we use the flow to warp the second segmentation mask $M_t$ with nearest sampling to obtain a synthesized mask of the first frame $\hat{M}_{t-1}(p) = M_t(p + V(p))$, where $p$ are the pixel coordinates.
The overlap of the warped mask $\hat{M}_{t-1}$ with the ground truth mask of the first frame $M_{t-1}$ gives us an estimate of the flow accuracy for objects in the scene,
and we use mean IoU as the metric.

\parsection{Data Deduplication.}
We found there is an overlap of 454 images between the segmentation training set and the detection and tracking validation image sets.
To maintain consistency in evaluation, we remove the overlapping images from the segmentation training set.
Single-task baselines are still trained with the full training set.
We found this to not noticeably affect the final results.


\subsection{VTD Accuracy Metric} \label{sec:sup-metric}
We provide additional details and analysis regarding our \ourbenchmark Accuracy (\ourmetric) metric.

\parsection{Details.}
\ourmetric first uses standard deviation estimates $\sigma_t$ and scaling factors $s_t = 1 / \lceil 2 \sigma_t \rceil$ for each task $t$ to normalize sensitivities of each metric.
$\sigma_t$ is measured over single-task baselines each trained with eight different base networks (ResNet-50/101~\cite{he2016deep}, Swin-T/S/B~\cite{liu2021swin}, and ConvNeXt-T/S/B~\cite{liu2022convnet}) and
are provided in \tableautorefname~\ref{tab:sup-metric}.
By computing standard deviation over these baselines, we can get an estimate of how network performances vary across different architectures and capacity.
Pose estimation and semantic segmentation have large variations in performance across different base networks.
On the other hand, drivable area segmentation and optical flow estimation have smaller variances.
Based on $\sigma_t$, we compute scaling factors $s_t$ in order to scale each task accordingly.
Pose estimation and semantic segmentation have a lower $s_t$, as improvements in these tasks are less significant.
Drivable area and optical flow have larger $s_t$, as minor improvements are more significant.
Each task score is multiplied by its corresponding $s_t$ to obtain the final score.

\parsection{Analysis.}
We compare absolute performance differences between \ourmodelbase and single-task baselines (row 3) to $s_t$ scaled performance differences (last row).
We also provide \ourmetric metrics for each aspect, which are calculated in the same way but using $\Delta$ instead.
With absolute difference, performance gains and losses in instance segmentation and pose estimation are large in magnitude and thus dominate the localization performance.
On the other hand, \ourmetric scales down their values as they are more sensitive than other metrics, leading to more balanced scores across the board.
Note that since we normalize the final scores of each aspect to the range $[0, 100]$, the magnitude of each score does not matter.
Similarly, absolute difference of tasks with lower sensitivity (\ie, drivable area segmentation) will not reflect the significance of the improvements in performance.
Thus, we scale the scores of such tasks relatively more compared to other tasks.

\begin{figure*}[t]
  \centering
   \includegraphics[width=0.95\linewidth]{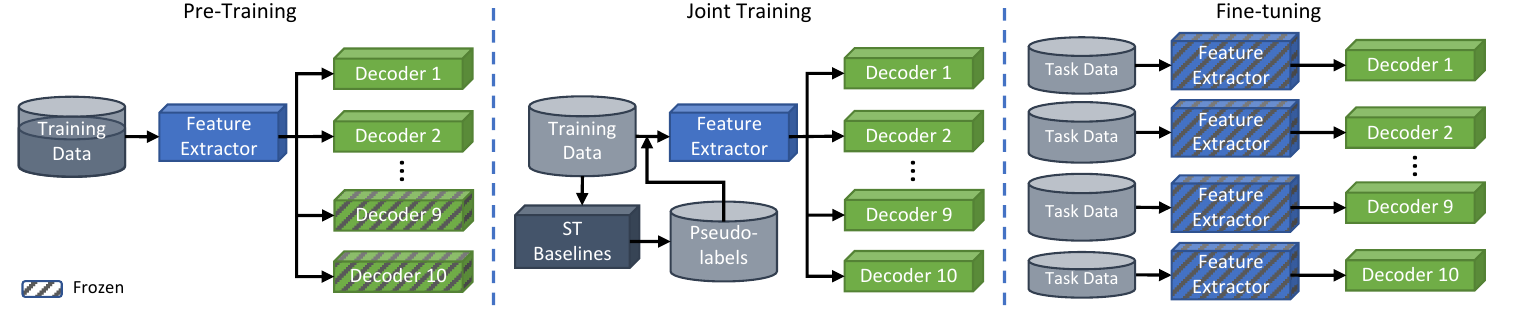}
    \vspace{-0.12in}
   \caption{
   Our training protocol, CPF, including Curriculum training, Pseudo-labeling, and Fine-tuning.
   A subset of the network is first pre-trained on a portion of data.
   Then, the network is jointly trained on all tasks, using pseudo-labels for label-deficient tasks to avoid undertraining.
   Finally, each decoder is independently fine-tuned on its respective data while freezing the learned shared representation.
   }
   \label{fig:cpf}
   \vspace{-0.12in}
\end{figure*}
\begin{figure*}[t]
\centering
\includegraphics[width=0.99\linewidth]{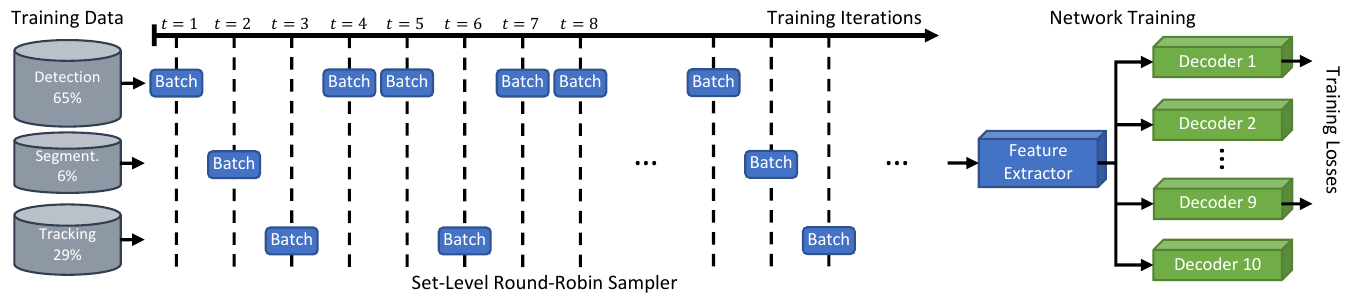}
\vspace{-0.16in}
\caption{
Illustration of our joint training protocol.
We use a set-level round-robin data sampler for sampling batches of data during training.
Each batch only contains annotations for a subset of the tasks, and the corresponding decoders are updated.
}
\label{fig:joint-training}
\vspace{-0.12in}
\end{figure*}

\section{\ourmodelbase Details} \label{sec:sup-hdet}
We provide additional details regarding task decoders and feature interaction blocks.

\subsection{Task Decoders}
We first describe details about certain decoders and loss functions used for training.


\parsection{Segmentation Decoders.}
The drivable area, lane detection, and semantic segmentation decoders use the same structure and employ convolutional layers to map the aggregated pixel features to the desired output.
The only exception is the lane detection decoder, as it requires making per-pixel predictions for three separate labels.
We replace the final convolutional layer with a convolutional layer for each label.
We replace all convolutions with deformable ones~\cite{zhu2019deformable} and use Group Normalization~\cite{wu2018group}.
For the lane detection decoder, we additionally scale the foreground pixels by 10 to better balance their loss against background pixels.
Cross entropy is used as the training loss for each decoder $L_\mathtt{S}$, $L_\mathtt{A}$, and $L_\mathtt{L}$.

\parsection{Localization Decoders.}
The training loss of the localization decoders $L_{\texttt{loc}}$ is a combination of multiple losses for the Region Proposal Network (RPN)~\cite{ren2015faster} $L_{\texttt{RPN}}$, bounding box decoder $L_{\texttt{D}}$, mask decoder $L_{\texttt{I}}$, and pose estimation decoder $L_{\texttt{P}}$,
\begin{align}
    L_\texttt{loc} = \lambda_\texttt{RPN} L_\texttt{RPN} + \lambda_\texttt{D} L_\texttt{D} + \lambda_\texttt{I} L_\texttt{I} + \lambda_\texttt{P} L_\texttt{P},
\end{align}
following Mask R-CNN~\cite{he2017mask}.
$L_{\texttt{RPN}}$ is a combination of a cross entropy loss for the classification branch of the RPN and a L1 loss for the regression branch.
Similarly, $L_{\texttt{D}}$ use the same losses for classification and regression.
$L_{\texttt{I}}$ uses a cross entropy loss on the instance mask predictions.
For the pose estimation decoder, instead of a one-hot heatmap indicating the location of the joint in the Region of Interest (RoI), we use a Gaussian distribution as the training targets, following~\cite{Xiao_2018_ECCV}.
The keypoint localization loss $L_{\texttt{P}}$ is then Mean Squared Error (MSE) on the predicted joint heatmaps.
We use $\lambda_{\texttt{RPN}}=1.0$ in our experiments by default.

\setlength{\tabcolsep}{6.5pt}
\begin{table*}[]
\footnotesize
\centering
\caption{
Training details of every single-task baseline and \ourmodelbase using ResNet-50.
}
\label{tab:sup-train-params}
\vspace{-0.14in}
\begin{tabular}{l|cccccc}
\toprule
Model & lr & Optimzer & Batch Size & Epochs & Schedule & Augmentations \\ \midrule
Image Tagging & 0.1 & SGD & 48 & 60 & Step decay at $[30, 45]$ & Random crop and flip \\
\midrule
Object Detection & 0.04 & \multirow{3}{*}{SGD} & 32 & \multirow{3}{*}{36} & \multirow{3}{*}{Step decay at $[24, 33]$} & \multirow{3}{*}{Multi-scale, random flip} \\
Instance Seg. & 0.02 &  & 16 &  &  &  \\
Pose Estimation & 0.02 &  & 16 &  &  &  \\
\midrule
Drivable Area & \multirow{3}{*}{0.01} & \multirow{3}{*}{SGD} & \multirow{3}{*}{16} & $\sim$18 (80K iters.) & \multirow{3}{*}{\begin{tabular}{@{}c@{}}Poly. decay with power $=0.9$,\\min. lr $=0.0001$\end{tabular}} & \multirow{3}{*}{\begin{tabular}{@{}c@{}}Random scale, crop, and flip;\\photometric distortion\end{tabular}} \\
Lane Detection &  &  &  & $\sim$18 (80K iters.) &  &  \\
Semantic Seg. &  &  &  & $\sim$183 (80K iters.) &  &  \\
\midrule
MOT & 0.02 & SGD & \multirow{3}{*}{16} & 12 & Step decay at $[8, 11]$ & Random flip \\
MOTS & 0.01 & SGD &  & 12 & Step decay at $[8, 11]$ & Random flip \\
Optical Flow & 0.0001 & AdamW &  & 36 & Step decay at $[24, 33]$ & Multi-scale, random flip \\
\midrule
\rowcolor{LightCyan} \textbf{\ourmodelbase} & 0.0001 & AdamW & 16 & 12 & Step decay at $[8, 11]$ & Multi-scale, random flip \\
\bottomrule
\end{tabular}
\vspace{-0.12in}
\end{table*}

\parsection{Association Decoders.}
The architecture of the optical flow estimation decoder follows PWCNet~\cite{sun2018pwc}.
The flow decoder uses the first two pixel feature maps from the Feature Pyramid Network (FPN)~\cite{lin2017feature} to create a feature pyramid.
At each pyramid level, features of the second image are warped using the upsampled flow from the previous layer and then used to compute a cost volume through the correlation operation.
Then, convolutional layers are used to predict the flow.
We reduce the number of parameters in PWCNet by reducing the number of dense connections in the flow decoder and only using four pyramid levels.
For occlusion estimation, we use the range map approach~\cite{wang2018occlusion}, which we found to greatly stabilize training.

The training loss of the flow decoder is a combination of a photometric consistency loss $L_{\texttt{photo}}$ and a smoothness constraint loss $L_{\texttt{smooth}}$, which are commonly used by unsupervised optical flow estimation methods,
\begin{align}
    L_\texttt{F} = \lambda_\texttt{photo} L_\texttt{photo} + \lambda_\texttt{smooth} L_\texttt{smooth}.
\end{align}
We use the Census loss~\cite{meister2018unflow} as $L_{\texttt{photo}}$ and the edge-aware second order smoothness constraint~\cite{tomasi1998bilateral} as $L_{\texttt{smooth}}$ with an edge-weight of $150.0$.
We use $\lambda_{\texttt{photo}}=1.0$ and $\lambda_{\texttt{smooth}}=4.0$ in our experiments by default.
We also experimented with using object segmentation masks as an additional training signal by enforcing consistency between the warped masks (similar to the evaluation protocol), but did not find it to be beneficial for performance.

The training loss of the MOT decoder is a combination of a multi-positive cross entropy loss $L_{\texttt{embed}}$ and a L2 auxiliary loss $L_{\texttt{aux}}$,
\begin{align}
    L_\texttt{T} = \lambda_{\texttt{embed}} L_{\texttt{embed}} + \lambda_{\texttt{aux}} L_{\texttt{aux}},
\end{align}
following QDTrack~\cite{pang2021quasi, fischer2022qdtrack}.
QDTrack uses an additional lightweight embedding head to extract features for each RoI from the RPN.
Contrastive learning is used on the dense RoIs of two video frames (key and reference frames) for feature learning. $L_{\texttt{embed}}$ is defined as,
\begin{align}
    L_\mathtt{\texttt{embed}} = \log \left[1 + \sum_{\boldsymbol{k}^+}\sum_{\boldsymbol{k}^-} \exp \left( \boldsymbol{v} \cdot \boldsymbol{k}^- - \boldsymbol{v} \cdot \boldsymbol{k}^+ \right) \right],
\end{align}
where $\boldsymbol{v}$ is the feature embeddings of the training sample in the key frame and $\boldsymbol{k}^+$ and $\boldsymbol{k}^-$ are its positive and negative targets in the reference frame. The auxiliary loss is used to constrain the magnitude and cosine similarity of the vectors, which is defined as
\begin{align}
    L_\mathtt{\texttt{aux}} = \log \left( \frac{\boldsymbol{v}\cdot\boldsymbol{k}}{||\boldsymbol{v}|| \cdot ||\boldsymbol{k}||} - c \right) ^2,
\end{align}
where $c=1$ if it is a positive match and $c=0$ otherwise.
We use $\lambda_{\texttt{embed}}=0.25$ and $\lambda_{\texttt{aux}}=1.0$ in our experiments by default.

For MOTS, we simply combine the outputs from the MOT and instance segmentation decoders, so there are no trainable parameters.

\subsection{Feature Interaction Blocks}
\ourmodel utilizes two types of feature sharing modules, Intra-group (Intra-IB) and Cross-group (Cross-IB) Interaction Blocks.
We use these blocks for the segmentation and localization task groups, but 
not the classification group as we found it does not require additional shared processing.

\parsection{Intra-IB}
uses self-attention blocks to model feature interactions within a task group.
For the segmentation task group, we use Window and Shifted Window Multi-Head Attention~\cite{liu2021swin} on the high resolution feature maps to reduce computation costs.
For the localization task group, we use the standard Multi-Head Attention~\cite{vaswani2017attention} on the object features.

\parsection{Cross-IB}
uses cross-attention blocks to model feature interactions between task groups.
However, such attention is expensive as the resolution of pixel features is high and the number of object features can be high during training.
To reduce the computational costs, we downsample the pixel features by a factor of 8 and average pool the object features into 1D vectors.

\section{\ourlearn Training Protocol Details} \label{sec:sup-cpf}
We provide additional details regarding each aspect of our training protocol \ourlearn.
The full protocol is illustrated in \figureautorefname~\ref{fig:cpf}.

\subsection{Curriculum Training} \label{sec:sup-curr}
Curriculum training involves pre-training a subset of the network first then joint-training the entire network.

\parsection{Pre-Training.}
We first train the feature extractor and localization and object tracking decoders on all three image sets.
This includes the object detection, instance segmentation, pose estimation, MOT, and MOTS tasks.
We follow the training procedure of QDTrack-MOTS~\cite{pang2021quasi, fischer2022qdtrack}, which first trains QDTrack on the detection and tracking sets then finetunes a instance segmentation decoder on the segmentation set and MOTS subset while freezing the rest of the network.
We additionally train the pose estimation decoder along with the instance segmentation decoder.

\parsection{Joint Training.}
We provide a detailed illustration of our joint training protocol in \figureautorefname~\ref{fig:joint-training}.
We use a set-level round-robin data sampler, which samples a batch of training data from each image set in order.
By default, we do not oversample the data in each set and spread out the samples to avoid many training iterations without gradients for a particular group of tasks.
For tracking, we use the MOTS subset instead of the full tracking set for better data balance, which is only 10\% the size.
This does not compromise MOT performance as we already pre-trained the MOT decoder on the full tracking set.
The loss weights used for joint training is provided in \tableautorefname~\ref{tab:sup-lossw} under the default setting.
The MOTS decoder has no trainable parameters, so there is no corresponding loss weight.

\begin{figure*}[t]
\centering
\includegraphics[width=\linewidth]{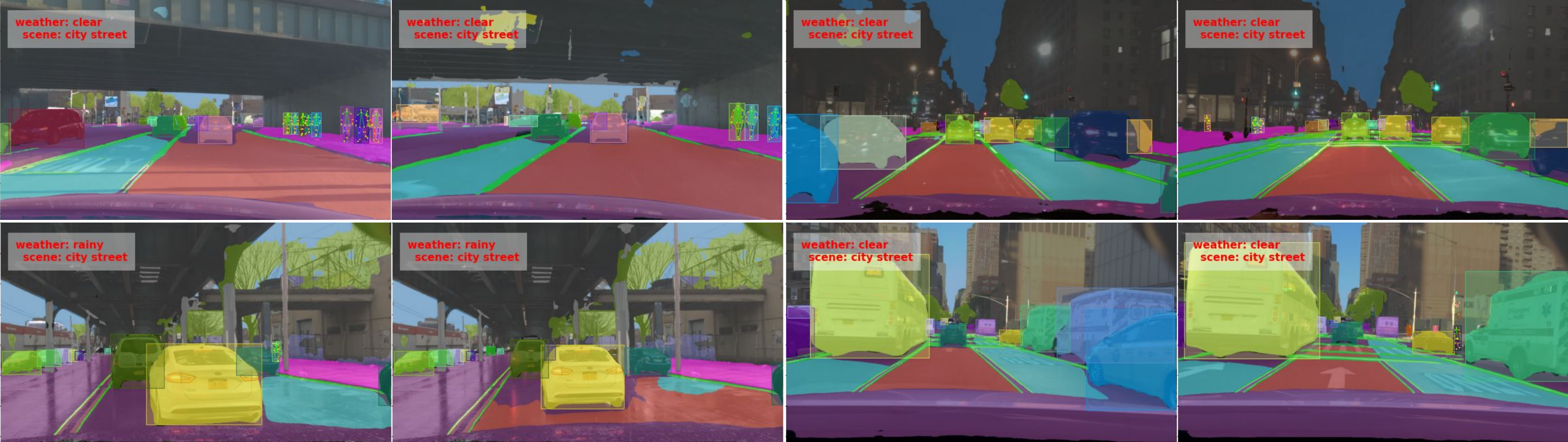}
\vspace{-0.28in}
\caption{
Additional visualizations of \ourmodelbase predictions on all tasks (excluding flow).
Best viewed in color.
}
\label{fig:sup-vis}
\vspace{-0.12in}
\end{figure*}

\begin{figure*}[t]
\centering
\includegraphics[width=\linewidth]{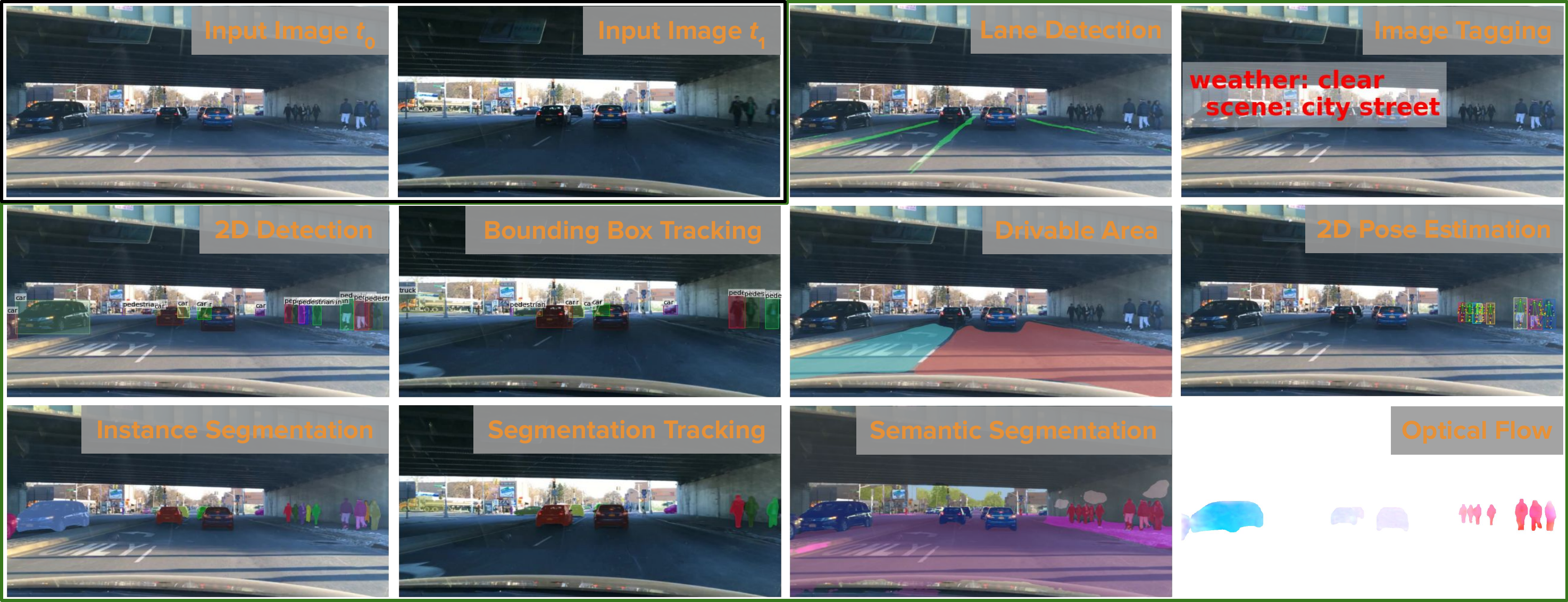}
\vspace{-0.28in}
\caption{
Visualization of \ourmodelbase predictions on all ten \ourbenchmark tasks for a pair of input images.
Best viewed in color.
}
\label{fig:sup-vis-tasks}
\vspace{-0.12in}
\end{figure*}

\subsection{Pseudo-Labeling}
We use single-task baselines to generate pseudo-labels for \ourmodelbase.
For consistency, we use single-task baselines with the same base network as \ourmodelbase to generate the pseudo-labels.
Such pseudo-labels are only used for pose estimation and semantic segmentation as the proportion of their data is the lowest.

\parsection{Pose Estimation.}
We generate pose estimation pseudo-labels following standard inference procedure.
We use a visibility threshold of 0.2 to filter the predictions, where predicted joints with a confidence lower than this threshold are removed.

\parsection{Semantic Segmentation.}
We generate semantic segmentation pseudo-labels using standard inference procedure.
We use a confidence threshold of 0.3 to filter the predictions,
where prediction pixels with a confidence lower than this threshold are set to unknown and ignored.

\subsection{Fine-Tuning}
After joint training, we fine-tune each task-specific decoder on the corresponding data for six epochs while freezing the rest of the network.
We decrease the learning rate by a factor of 10.
After fine-tuning, we combine the weights of each decoder and the rest of the network to obtain our final network weights.

\section{Training Details} \label{sec:sup-tdet}
In this section, we show full training details for~\ourmodelbase and the single-task baselines.
All models are trained on either 8 GeForce RTX 2080 Ti or 8 GeForce RTX 3090 GPUs.
We use half-precision floating-point format (FP16) for all models to speed up training.
We use the same codebase and environment for training all models to ensure consistency in the training setting.
For each task, the baseline is trained only on data from the particular task.
The baseline uses task-specific augmentations and training schedules that are optimized for single-task performance, which we detail in \tableautorefname~\ref{tab:sup-train-params}.
For SGD~\cite{SGD}, we use a momentum of 0.9 and weight decay of $10^{-4}$.
For AdamW~\cite{kingma2014adam, loshchilov2017decoupled}, we use $\beta_1=0.9$, $\beta_2=0.999$, and weight decay of 0.05.
For the multi-scale augmentation, we sample an image height from [$600, 624, 648, 672, 696, 720$] and scale the image while keeping the aspect ratio the same.
For all models using Swin Transformer~\cite{liu2021swin} or ConvNeXt~\cite{liu2022convnet} as the base network, we use AdamW with a learning rate of 0.0001.

\section{Visualizations} \label{sec:sup-vis}
We provide additional visualizations of~\ourmodelbase predictions on the~\ourbenchmark tasks in Figure~\ref{fig:sup-vis}.
We also visualize each task prediction separately in Figure~\ref{fig:sup-vis-tasks}.
The color of each object indicate the predicted instance identity.
For drivable area segmentation, red areas on the road indicate drivable regions, and blue areas indicate alternatively drivable areas.
The green lines represent predicted lanes on the road.
For optical flow estimation, we segment the flow using the instance segmentation mask predictions to extract object-level flow to be consistent with the evaluation protocol.
\ourmodelbase can produce high quality predictions for all ten tasks.


\end{document}